# A Nonlinear African Vulture Optimization Algorithm Combining Henon Chaotic Mapping Theory and Reverse Learning Competition Strategy

Baiyi Wang[1], Zipeng Zhang[1], Patrick Siarry[2], Xinhua Liu[1], Grzegorz Królczyk[3], Dezheng Hua[1], Frantisek Brumercik[4], Zhixiong Li[5*]

1. School of Mechatronic Engineering, China University of Mining and Technology, Xuzhou, China
2. University Paris-Est Créteil Val de Marne, 61 Av. du General de Gaulle, 94010, Creteil, France
3. Department of Manufacturing Engineering and Automation Products, Opole University of Technology, 45-758 Opole, Poland
4. Department of Design and Machine Elements, Faculty of Mechanical Engineering, University of Zilina, Univerzitna 1, 010 26 Zilina, Slovakia
5. Yonsei Frontier Lab, Yonsei University, Seoul 03722, South Korea

Emails: ts22050054a31ld@cumt.edu.cn (Baiyi Wang); cnzzp@cumt.edu.cn (Zipeng Zhang); siarry@u-pec.fr (Patrick Siarry); liuxinhua@cumt.edu.cn (Xinhua Liu); g.krolczyk@po.edu.pl (Grzegorz Królczyk); hua_dezheng@cumt.edu.cn (Dezheng Hua); frantisek.brumercik@fstroj.uniza.sk (Frantisek Brumercik)

* Corresponding author: zhixiong.li@yonsei.ac.kr (Zhixiong Li)

**Abstract**: As a new intelligent optimization algorithm, the African vultures optimization algorithm (AVOA) has been widely used in various fields today. However, when solving complex multimodal problems, the AVOA still has some shortcomings, such as low searching accuracy, deficiency on the search capability and tendency to fall into local optimum. In order to alleviate the main shortcomings of the AVOA, a nonlinear African vulture optimization algorithm combining Henon chaotic mapping theory and reverse learning competition strategy (HWEAVOA) is proposed. Firstly, the Henon chaotic mapping theory and elite population strategy are proposed to improve the randomness and diversity of the vulture's initial population; Furthermore, the nonlinear adaptive incremental inertial weight factor is introduced in the location update phase to rationally balance the exploration and exploitation abilities, and avoid individual falling into a local optimum; The reverse learning competition strategy is designed to expand the discovery fields for the optimal solution and strengthen the ability to jump out of the local optimal solution. HWEAVOA and other advanced comparison algorithms are used to solve classical and CEC2022 test functions. Compared with other algorithms, the convergence curves of the HWEAVOA drop faster and the line bodies are smoother. These experimental results show the proposed HWEAVOA is ranked first in all test functions, which is superior to the comparison algorithms in convergence speed, optimization ability, and solution stability. Meanwhile, HWEAVOA has reached the general level in the algorithm complexity, and its overall performance is competitive in the swarm intelligence algorithms.

## 1. Introduction

Optimization problems are common in many fields, such as intelligent production, scientific research, and economic management. Today, the complexity and difficulty of optimizing issues are increasing, and solutions are becoming more dynamic and computationally complex. Finding one or more points in a multidimensional hyperspace is often necessary. Traditional data processing methods are increasingly difficult to cope with the data surge problems brought at the digital age. Therefore, intelligent optimization algorithms are needed to determine an accurate solution for us (Kar, 2016; D. Karaboga & Akay, 2009; Li, Wang, & Gandomi, 2021). The intelligent optimization algorithm is a new optimization algorithm that simulates biological behaviours and some natural physical phenomena. With the appliance of these algorithms, many complex optimization problems can be solved efficiently (Valdez, Castillo, Cortes-antonio, & Melin, 2022). Compared with a traditional optimization algorithm, an intelligent optimization algorithm converges fast, robust, pervasive and stable (Cui, Geng, Zhu, & Han, 2017; Nabaei, et al., 2018). In recent years, intelligent optimization algorithms have developed unprecedently. Scholars put forward a series of new intelligent optimization algorithms, such as the supply-demand-based optimization (Zhao, Wang, & Zhang, 2019), carnivorous plant algorithm (Meng, Pauline, & Kiong, 2021), honey badger optimization algorithm (Hashim, Houssein, Hussain, Mabrouk, & Al-Atabany, 2022), Runge Kutta optimization algorithm (Ahmadianfar, Heidari, Gandomi, Chu, & Chen, 2021), hunger game search algorithm (Yang, Chen, Heidari, & Gandomi, 2021), wild horse optimization algorithm (Naruei & Keynia, 2022), material generation optimization algorithm (Oyelade, Ezugwu, Mohamed, & Abualigah, 2022), spider jumping optimization algorithm (Peraza-Vazquez, et al., 2022), reptile search algorithm (Abualigah, Abd Elaziz, Sumari, Geem, & Gandomi, 2022), capuchin search algorithm (Braik, Sheta,

& Al-Hiary, 2021). The intelligent optimization is widely used in system identification (N. Karaboga, 2009), path planning (Wang, Yan, & Gu, 2019), troubleshooting (Deng, Li, Li, Chen, & Zhao, 2022), neural networks (Xu, Yu, & Gulliver, 2021), optimization control (Hamza, Yap, & Choudhury, 2017) and other fields (Kalinli & Karaboga, 2005; N. Karaboga & Cetinkaya, 2011) for good searchability.

The African vultures optimization algorithm (AVOA) (Abdollahzadeh, Gharehchopogh, & Mirjalili, 2021) is a new intelligent optimization algorithm proposed by Benyamin Abdollahzadeh et al. in 2021, which simulates the foraging and navigation behaviours of African vultures. The algorithm has the advantages of simple structure, easy implementation and outstanding performance in finding optimal values, which has been well applied in various fields. (Salah, et al., 2022) introduced the African vultures optimization algorithm to optimize the PID controller and apply it to control DC microgrid voltage. (Mekala, Sumathi, & Shobana, 2022) proposed a multi-polymer charging scheduling strategy based on AVOA to realize the proper planning of electric vehicle charging. (Singh, Houssein, Mirjalili, Cao, & Selvachandran, 2022) introduced the African vulture algorithm to optimize solutions to the travel salesman shortest path problem. (Diab, Tolba, El-Rifaie, & Denis, 2022) introduced the African vulture algorithm to accurately predict unknown parameters of various solar photovoltaic units. (Zhang, Khayatnezhad, & Ghadimi, 2022) applied the African vulture algorithm to the actual PEMFC baseline case study and established an optimal evaluation model for fuel cells.

Although the AVOA has a large amount of applications, it still shows deficiency on the search capability and tendency to fall into local optimum. To solve these problems of the AVOA, (Liu, et al., 2022) introduced quasi-antagonistic learning mechanisms, differential evolution operators and adaptive parameters to balance AVOA exploration and development capabilities. (Fan, Li, & Wang, 2021) uses chaotic mapping and time-varying tool to optimize the global optimal solution and convergence performance of AVOA. (Soliman, Hasanien, Turky, & Muyeen, 2022) proposed a new African vulture-grey wolf hybrid optimizer to improve the convergence speed and stability of the algorithm. (Kannan, Mannathazhathu, & Raghavan, 2022) proposed a hybrid optimization algorithm based on honey badger and African vulture, the global optimization search of the algorithm is implemented, and the probability of falling into the local optimum is reduced.

To date, due to the novelty, there are few studies on the improvements of AVOA. Although the existing improved AVOA increases the optimization performance, there are still some limitations and uncertainties, such as the lack of diversity in initialized populations, the unbalance of the exploration and exploitation capabilities, and the waste of valuable population information, which resulting in the AVOA algorithm is sensitive to local optimal, cannot get the ideal solution. Therefore, an improved AVOA with multi-strategy (HWEAVOA) is proposed to eliminate the uncertainty and restriction of the original AVOA, and subsequently improve its algorithm's performance. Firstly, Henon chaotic mapping theory and elite population strategy (HCE) are proposed, which make the initial population distribution more homogeneous, enhance the global optimization performance and the convergence rate of the AVOA. Then, the nonlinear adaptive incremental inertial weight factor (NWF) is introduced to optimally update the position of vultures. This strategy can assist vulture populations to search at different convergence rates, which balances the exploration and exploitation abilities, and effectively avoids AVOA falling into a local optimum. Finally, the reverse learning competition strategy (RLC) is designed to increase the diversity of the population. The bad performed vulture individuals are given learning opportunities, and they will have the probability to become dominant individuals. This strategy expands the discovery fields for the optimal solution and avoids the generation of local optimum phenomenon.

To evaluate the effectiveness of the HWEAVOA algorithm, classical and CEC2022 test functions are used to compare the optimization performance of the AVOA and other improved algorithms.

The rest of this paper is organized as follows. In Section 2, the original AVOA is briefly introduced. In section 3, the technical details of the HWEAVOA algorithm are described. In section 4, the performance of HWEAVOA is analyzed reliably by using classical and CEC2022 test functions. In section 5, this study is summarized by discussing the results and possible future areas for potential investigations are puts forward.

## 2. African vultures optimization algorithm (AVOA)

The African vultures optimization algorithm (AVOA) simulates the foraging and navigation behaviours of African vultures from the African vulture lifestyle. Each individual in the population relies on their hunger rate for corresponding behaviours and completes the switch between the exploration and development stages. The hunger rate is calculated as follows:

$$M = h \times \left( sin^W \left( \frac{\pi}{2} \times \frac{t}{T} \right) + cos \left( \frac{\pi}{2} \times \frac{t}{T} \right) - 1 \right) \qquad (1)$$

$$F = (2 \times rand_1 + 1) \times z \times \left( 1 - \frac{t}{T} \right) + M \qquad (2)$$

Where: $F$ is the vulture hunger rate, $t$ indicates the current number of iterations, $T$ is the maximum number of iterations, $W$ shows a fixed parameter set before the algorithm works, $z$ represents a random number between -1 and 1, $h$ represents a random number between -2 and 2, and $rand_1$ indicates a random number between 0 and 1.

When the $F$ value is greater than 1, the vulture searches for food in different regions and enters the exploration phase, using formula (3) to search for food in other areas.

$$P(i+1) = \begin{cases} \text{Optimal solution guidance strategy, if } P_1 \geq rand_{P_1} \\ \text{Random search strategy, if } P_1 < rand_{P_1} \end{cases} \qquad (3)$$

Where: $P_1$ is a control parameter with values between 0 and 1; $rand_{P_1}$ represents a random number between 0 and 1.

About the optimal solution guidance strategy, the remaining vultures search for food near one of the optimal vultures at a random distance. The position update formula is as follows:

$$P(i+1) = R(i) - |X \times R(i) - P(i)| \times F \qquad (4)$$

Where: $P(i+1)$ represents the vulture position vector in the next iteration, $F$ is the hunger rate of vulture individuals in the current iteration. $X$ is a place where vultures move randomly, which is used as a coefficient vector to increase random motion and obtained by using the formula $X = 2rand$, where $rand$ is a random number between 0 and 1; $P(i)$ indicates the current vector position of the vulture. $R(i)$ indicates the best vulture chosen at random, and the solution formula is as follows:

$$R(i) = \begin{cases} BestV_1, p_i = L_1 \\ BestV_2, p_i = L_2 \end{cases} \qquad (5)$$

Where $BestV_1$, $BestV_2$ represents the two best adapted vultures in the vulture population respectively; $L_1$, $L_2$ represents the parameters between 0 ~ 1 which waited to be measured respectively, and their sum is 1; $p_i$ represents the probability of selecting the best vulture;

On the other hand, vultures perform random search strategies with the following positional update formula (6).

$$P(i+1) = R(i) - F + rand_2 \times ((ub - lb) \times rand_3 + lb) \quad (6)$$

Where $rand_2$ and $rand_3$ are random values between 0 and 1, $lb$ and $ub$ represent the upper and lower bounds of the variable.

If the value of $F$ is less than 1, the vulture enters the development phase and looks for food near the best solution. When $0.5 \leq |F| \leq 1$, as shown in Fig. 1, the population gets food through the implementation of the conflict profit strategy and rotational flight strategy, the two methods are selected and executed by formula (7).

$$P(i+1) = \begin{cases} \text{Conflict profit strategy}, & \text{if } P_2 \geq rand_{P_2} \\ \text{Rotational flight strategy}, & \text{if } P_2 < rand_{P_2} \end{cases} \quad (7)$$

where $P_2$ is a control parameter with values between 0 and 1; $rand_{P_2}$ is a random number between 0 and 1.

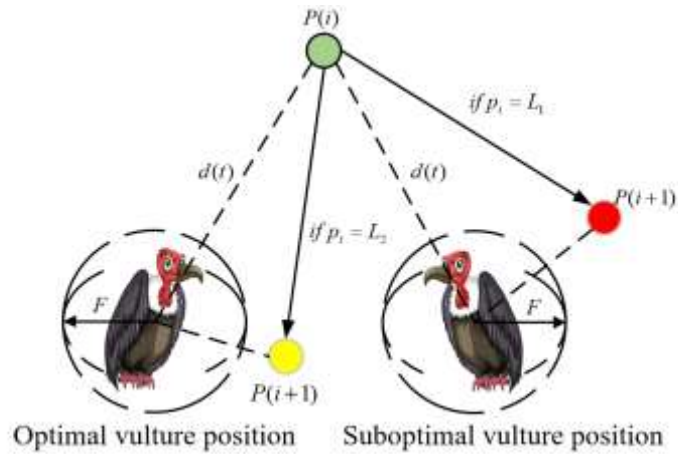

Fig. 1. The schematic map of vulture populations vying for food.

About the conflict profitability strategy, weak vultures try to get food by causing conflicts between healthy vultures to fatigue them, and their position update formula is updated as follows:

$$P(i+1) = |X \times R(i) - P(i)| \times (F + rand_4) - (R(i) - P(i)) \tag{8}$$

where $rand_4$ is a random number between 0 and 1.

In addition, the vulture rotational flight strategy is as follows:

$$\begin{cases} S_1 = R(i) \times \left( \dfrac{rand_5 \times P(i)}{2\pi} \right) \times \cos(P(i)) \\ S_2 = R(i) \times \left( \dfrac{rand_6 \times P(i)}{2\pi} \right) \times \sin(P(i)) \end{cases} \tag{9}$$

$$P(i+1) = R(i) - (S_1 + S_2) \tag{10}$$

where $rand_5$ and $rand_6$ are random numbers between 0 and 1; $S_1$ and $S_2$ are calculated by formula (8). Finally, the vulture position update is completed by formula (9).

When $|F| \leq 0.5$, as shown in Fig. 2, the population gets food through the implementation of individual competition strategy and population competition strategy, two methods are selected and executed by formula (11).

$$P(i+1) = \begin{cases} \text{Individual competition strategy}, if\ P_3 \geq rand_{P_3} \\ \text{Population competition strategy}, if\ P_3 < rand_{P_3} \end{cases} \tag{11}$$

where $P_3$ is a control parameter with values between 0 and 1; $rand_{P_3}$ is a random number between 0 and 1.

When an individual competition strategy is implemented, multiple vultures may accumulate on the same food source, and the position update formula is as follows:

$$\begin{cases} A_1 = BestV_1 - \dfrac{BestV_1 \times P(i)}{BestV_1 - P(i)^2} \times F \\ A_2 = BestV_2 - \dfrac{BestV_2 \times P(i)}{BestV_2 - P(i)^2} \times F \end{cases} \tag{12}$$

$$P(i+1) = \dfrac{A_1 + A_2}{2} \tag{13}$$

When a population competition strategy is implemented, multiple vultures may accumulate on the same food source, and the position update formula is as follows:

$$P(i+1) = R(i) - |R(i) - P(i)| \times F \times Levy \tag{14}$$

where $Levy$ indicates the levy flight.

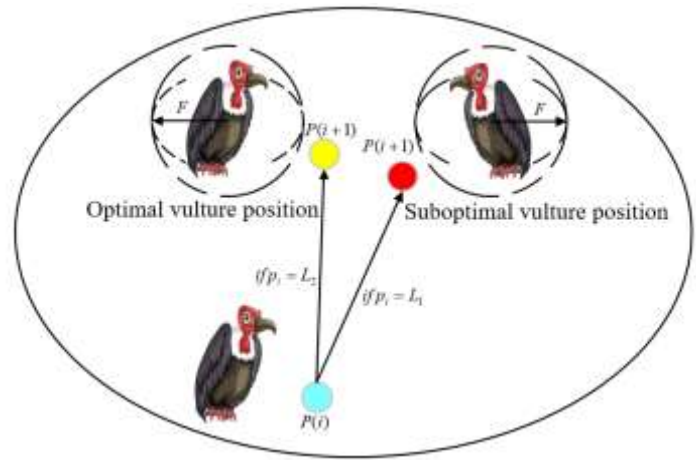

Fig. 2. The schematic map of vulture individuals fiercely competing for food.

## 3. The proposed optimization algorithm (HWEAVOA)

As mentioned above, although the overall mechanism of the AVOA algorithm is simple and easy to implement, it still has some limitations, such as the lack of diversity in initialized populations, the unbalance of the exploration and exploitation capabilities, and the waste of valuable population information, which resulting in the AVOA algorithm is sensitive to local optimal, cannot get the ideal solution. In this section, three improvement strategies are proposed, namely the Henon chaotic mapping theory and elite population strategy (HCE), the nonlinear adaptive incremental inertial weight factor (NWF), and the reverse learning competition strategy (RLC), which will be discussed in the following sections.

### 3.1 Henon chaotic mapping theory and elite population strategy (HCE)

The primitive vulture population is initialized by randomization, with an uneven distribution and lacking population. Cause of low randomness, primitive AVOA always faces high uncertainty. Therefore, if the homogeneous initialization of the population can be solved, the population diversity can be increased effectively, and the searching efficiency of the algorithm can be improved. To this end, Henon chaotic mapping theory and elite population strategy are introduced in this paper.

Henon chaotic mapping theory is a nonlinear theory with the characteristics of nonlinear, initial sensitivity, randomness and ergodicity. Henon chaotic map is produced in 2-dimensional space, a typical discrete chaotic map. Its kinetic formula is as follows:

$$\begin{cases} x_{n+1} = 1 + y_n - ax_n^2 \\ y_{n+1} = bx_n \end{cases} \tag{15}$$

Where the state of the Henon chaotic map is determined by the four parameters $x_0$, $y_0$, $a$, and $b$, which is more complex than the 1-dimensional chaotic map. This paper takes $a = 1.4$, $b = 0.3$ to ensure the strong randomness of the generated chaotic sequence when the function enters the chaotic state. By the above mapping method, the chaotic mapping initialization population is obtained.

Then, the elite population strategy is adopted by this paper, which combines the chaotic mapped initialized population and the conventional initialized population, calculates the adaptability of each initial vulture, and sorts it. At last, the first N elite individuals are selected, and the sequence of privileged individuals is as follows:

$$x_i = \{x_1, x_2, \ldots, x_N\}, i = 1 \square N \tag{16}$$

Where $x_1 = BestV_1$, $x_2 = BestV_2$. $N$ is the number of vultures in the population.

The above improved ways make the initial population distribution more homogeneous and give the initial population more possibilities, which enhances the global optimization performance and convergence rate of the AVOA.

**3.2 Nonlinear adaptive incremental inertial weight factor (NWF)**

In AVOA, the entire vulture population starts from the global search and gradually returns to local search, which performs mechanically the exploration phase and exploitation phase under the leadership of the optimal vulture and the suboptimal vulture. However, the whole process of algorithm is not static. It is difficult to effectively balance the global search stage and the local search stage of the population by using the original position update formula of AVOA. The entire process ignores the actual environment of the vulture population, which cause the AVOA is sluggish in convergence and prone to the local optimum. Therefore, the nonlinear adaptive incremental inertial weight factor is introduced to realize the rational allocation of the global exploration phase and local exploitation phase in different evolutionary periods in this paper.

The nonlinear adaptive incremental inertial weight factor $\omega$ is added to the process of the position renewal of vulture populations, which is calculated as follows:

$$\omega = \begin{cases} (\alpha + \beta \times rand) \times sin(\frac{\pi}{10} \times \frac{t}{T})^5, & mod(\frac{t}{6}) \geq 3 \\ 1, & mod(\frac{t}{6}) \leq 3 \end{cases} \quad (17)$$

where $\alpha$ and $\beta$ are the selection factors for the initial optimal vulture and the secondary vulture, and $rand$ is the random number of $0 \sim 1$. $mod(\frac{t}{6})$ is the residual number of iterations divided by 6.

The nonlinear adaptive incremental inertial weight factor $\omega$ is then introduced into the vulture position update formulas in the exploration and development phases; the process is shown in formulas (18), (19) and (20).

$$P(i+1) = \begin{cases} \omega \times Equation(4) & if \ P_1 \geq rand_{P_1} \\ \omega \times Equation(6) & if \ P_1 < rand_{P_1} \end{cases} \quad (18)$$

$$P(i+1) = \begin{cases} \omega \times Equation(8) & if \ P_2 \geq rand_{P_2} \\ \omega \times Equation(10) & if \ P_2 < rand_{P_2} \end{cases} \quad (19)$$

$$P(i+1) = \begin{cases} \omega \times Equation(13) & if \ P_3 \geq rand_{P_3} \\ \omega \times Equation(14) & if \ P_3 < rand_{P_3} \end{cases} \quad (20)$$

The position of vultures is optimally updated through the above formula. This step considers evolutionary differences between population vultures during evolution, which adaptively confers inertial weight factors of different sizes. When the inertial weight factor $\omega$ increases, the global optimization ability of the algorithm is significantly enhanced. However, the local search ability is reduced, and the solution accuracy could be lower. When the inertial weight factor $\omega$ declines, the global optimization ability is decreased, while the local optimization ability is enhanced and the solution accuracy is higher. The inertial weight factor assists vultures in searching at different convergence rates until they approach the optimal value in the next iteration. The inertial weight factor improves the search ability of the vulture population at different stages on the premise that the overall search behavior is unchanged. In the early stage of the population with a strong global search ability, its local search ability is improved appropriately, which increasing the search accuracy and convergence rate of the AVOA. In the later stage of the population with a strong local search ability, its global search ability is improved appropriately, which effectively avoiding AVOA falling into a local optimum. These characteristics meet the needs of the AVOA for global exploration ability and local exploitation ability at different evolutionary times.

### 3.3 Reverse learning competition strategy (RLC)

The information from the optimal location in the iteration process is used simply in the original AVOA, and part of the valuable information in the population is wasted easily. Therefore, the reverse learning competition strategy is introduced in this paper. The main idea of reverse competitive learning strategy is to increase the diversity of the population and chance of obtaining better solutions by simultaneously exploring the positive and negative direction of search space simultaneously. As shown in Fig. 3, the bad performed vulture individuals are given more learning opportunities, and they will have the probability to become dominant individuals. In this way, the loss of useful information is solved well and the accuracy of the next input is effectively guaranteed.

Based on each output solution, the reverse learning solution $E_P(i+1)$ is obtained through the reverse learning competition strategy. The calculation formula is as follows:

$$E_P(i+1) = rand \times (ub + lb) - P(i+1) \qquad (21)$$

where $rand$ is a random number of $0 \sim 1$.

The output position of vultures in this iteration is optimized by calculating the population fitness of $P(i+1)$, and $E_P(i+1)$. In this way, the optimal individual position is not lost, and the optimal individual information can be used to significantly improve the robustness of AVOA algorithm.

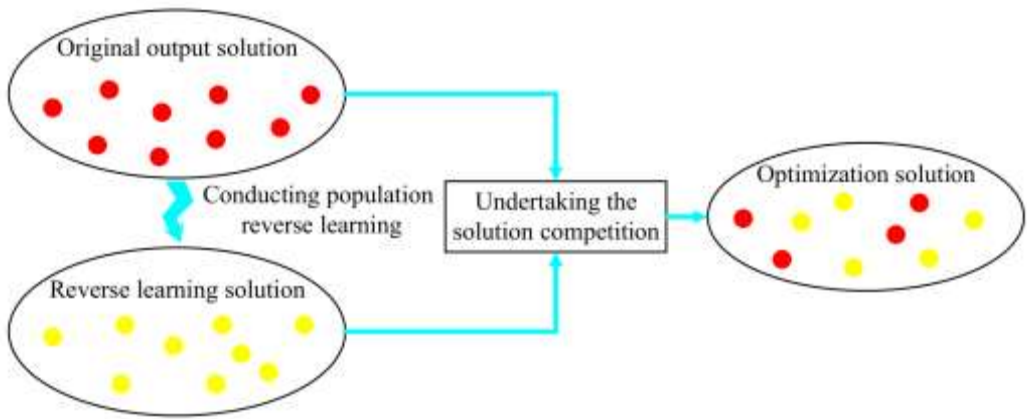

Fig. 3 Schematic map of reverse learning competition strategy.

In summary, the problems in the original AVOA algorithm, such as uneven population initialization individual distribution, lack of population diversity, no reasonable balance between global and local search stages, and easy loss of better personal information, are well solved based on three strategies, i.e., HCE, NWF, and RLC, corresponding to the Henon chaotic mapping theory and elite population strategy, the nonlinear adaptive incremental inertial weight factor, the reverse learning competition strategy, respectively. The framework of HWEAVOA algorithm is shown in the Fig. 4 and its technical details are described in Table 1.

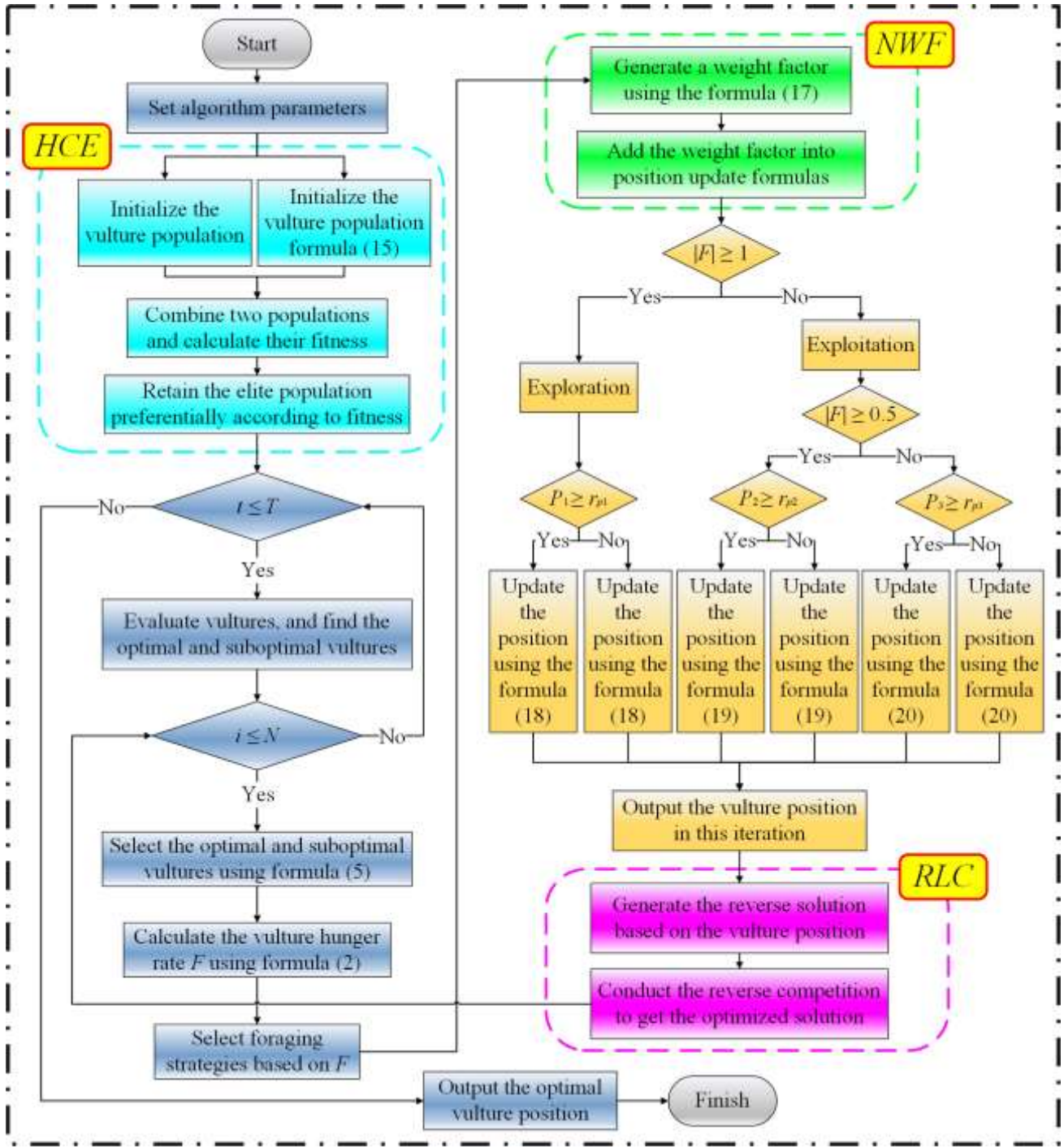

Fig. 4. The flowchart of HWEAVOA.

Table 1. The pseudo-code of HWEAVOA.

---

**Input:** The population size $N$, maximum number of iterations $T$, parameters used to determine each strategy's choosing $P_1$ $P_2$ $P_3$, the selection factors for the initial optimal vulture $a$ $β$, parameter with a fixed number $W$, Henon chaotic mapping parameters $a$ $b$
**Output:** The position of Vulture and its fitness value
Initialize the golden jackal population $P_i$ ($i$ = 1, 2, …, N)
    **Sub-Algorithm 1:** Henon chaotic mapping theory and elite population strategy
    Calculate the chaotic mapped initialized population using Eq (15)
    Combine the chaotic mapped initialized population and the conventional initialized population, calculate the fitness of vultures, and obtain the elite population
**while** $t$ =1 to $T$
    Set $BestV_1$ and $BestV_2$ as the position of the optimal vulture and suboptimal vulture
    **for** (each Vulture ($P_i$)) do
      Select $R(i)$ using Eq (5)
        Update the $F$ using Eq (2)
    **Sub-Algorithm 2:** Nonlinear adaptive incremental inertial weight factor
    Obtain the weight factor using Eq (17), introduced it into the position update formulas of vultures
    **if** ($|F| \geq 1$) (Exploration phase) then
      Update the vulture's location using Eq (18)
    **else** ($|F| < 1$) (Exploitation phase) then
      **if** ($|F| \geq 0.5$) then
        Update the vulture's location using Eq (19)
      **else** ($|F| < 0.5$) then
        Update the vulture's location using Eq (20)
      **end if**
    **end if**
    **Sub-Algorithm 3:** Reverse learning competition strategy
    Generate reverse learning solutions using Eq (21)
    Decide whether to update the location of the vulture according to the fitness of the reverse learning solutions and the original solutions
    **end for**
**end while**
Return $P_{BestV_1}$

---

## 4. Experimental results and discussion

Test experiments based on classical and CEC2022 test functions are carried out to validate the performance of the HWEAVOA in this section. The mathematical expressions and function characteristics of classical and CEC2022 test functions are given in Appendix 1 and Appendix 2 respectively.

All experiments were conducted over Windows 10 (64 bit) that runs on CPU Core i7 with 16GB RAM, and Matlab2021a are utilized. In order to minimize the randomness of the algorithm, each algorithm accumulates 1000 tests on the classical test functions, and the population size and maximum number of iterations for all algorithms are set to 30 and 500. In the CEC2022, each algorithm accumulates 30 tests, the population size is set to 20, and the maximum number of iterations for all algorithms is set to 200000 and 1,000,000 respectively in the condition of 10 and 20 dimensions.

The test functions are tested by these algorithms, and the search performance and optimize performance of each algorithm are detected by comparing the test results with the optimal value of the functions. The results mainly contain average (Avg) and standard deviation (Std), where the average (Avg) can verify the optimization ability of the algorithm, the standard deviation (Std) can verify the stability of the optimization process of the algorithm. The data shown in Table 2, Table 4, Table 5, Table 6, Table 7, Table 8, Table 9, Table 10, Table 11, Table 12, Table 13 and Table 14 are the results of the comparison optimization algorithms in solving these benchmark function problems.

In this section, HWEAVOA is compared with the original AVOA, 6 AVOA variant algorithms based on three improved strategies, 10 advanced intelligent optimization algorithms to verify that the better optimization performance of the HWEAVOA.

**4.1 Parameters tuning and analysis**

In HWEAVOA, $W$ (used in Eq. (1)) is mainly used to balance the exploration and development stages of vulture populations, $P_1$ (used in Eq. (18)), $P_2$ (used in Eq. (19)), and $P_3$ (used in Eq. (20)) are mainly used to assist vultures in selecting different behaviours and update their positions. In the original AVOA, the four parameters are derived from AVOA and are set as 2.5, 0.6, 0.4 and 0.6, respectively in AVOA. This section aims to study the effects of different parameters of the algorithm on the performance of HWEAVOA. The value of $W$ is increased from 2 to 3 with a step size of 0.5, the value of $P_1$, $P_2$, and $P_3$ are increased from 0.4 to 0.6 with a step size of 0.1. In this paper, the control variable method is adopted, only one parameter is changed in each experiment, and the values of other parameters are set according to the value of AVOA. For example, when the value of $W$ is increased from 2 to 3, the values of $P_1$, $P_2$, and $P_3$ is set as 0.6, 0.4 and 0.6; When the value of $P_1$ increases from 0.4 to 0.6, the values of $W$, $P_2$, and $P_3$ are set as 2.5, 0.4 and 0.6. When the value of $P_2$

Table 2. Effect of main parameters on HWEAVOA.

| | | $W=2$ | $W=2.5$ | $W=3$ | $P_1=0.4$ | $P_1=0.5$ | $P_1=0.6$ | $P_2=0.4$ | $P_2=0.5$ | $P_2=0.6$ | $P_3=0.4$ | $P_3=0.5$ | $P_3=0.6$ |
|---|---|---|---|---|---|---|---|---|---|---|---|---|---|
| F1 | Avg | 0.0000E+00 | 0.0000E+00 | 0.0000E+00 | 0.0000E+00 | 0.0000E+00 | 0.0000E+00 | 0.0000E+00 | 0.0000E+00 | 0.0000E+00 | 0.0000E+00 | 0.0000E+00 | 0.0000E+00 |
| | Std | 0.0000E+00 | 0.0000E+00 | 0.0000E+00 | 0.0000E+00 | 0.0000E+00 | 0.0000E+00 | 0.0000E+00 | 0.0000E+00 | 0.0000E+00 | 0.0000E+00 | 0.0000E+00 | 0.0000E+00 |
| F2 | Avg | 0.0000E+00 | 0.0000E+00 | 0.0000E+00 | 0.0000E+00 | 0.0000E+00 | 0.0000E+00 | 0.0000E+00 | 0.0000E+00 | 0.0000E+00 | 0.0000E+00 | 0.0000E+00 | 0.0000E+00 |
| | Std | 0.0000E+00 | 0.0000E+00 | 0.0000E+00 | 0.0000E+00 | 0.0000E+00 | 0.0000E+00 | 0.0000E+00 | 0.0000E+00 | 0.0000E+00 | 0.0000E+00 | 0.0000E+00 | 0.0000E+00 |
| F3 | Avg | 0.0000E+00 | 0.0000E+00 | 0.0000E+00 | 0.0000E+00 | 0.0000E+00 | 0.0000E+00 | 0.0000E+00 | 0.0000E+00 | 0.0000E+00 | 0.0000E+00 | 0.0000E+00 | 0.0000E+00 |
| | Std | 0.0000E+00 | 0.0000E+00 | 0.0000E+00 | 0.0000E+00 | 0.0000E+00 | 0.0000E+00 | 0.0000E+00 | 0.0000E+00 | 0.0000E+00 | 0.0000E+00 | 0.0000E+00 | 0.0000E+00 |
| F4 | Avg | 0.0000E+00 | 0.0000E+00 | 0.0000E+00 | 0.0000E+00 | 0.0000E+00 | 0.0000E+00 | 0.0000E+00 | 0.0000E+00 | 0.0000E+00 | 0.0000E+00 | 0.0000E+00 | 0.0000E+00 |
| | Std | 0.0000E+00 | 0.0000E+00 | 0.0000E+00 | 0.0000E+00 | 0.0000E+00 | 0.0000E+00 | 0.0000E+00 | 0.0000E+00 | 0.0000E+00 | 0.0000E+00 | 0.0000E+00 | 0.0000E+00 |
| F5 | Avg | 2.6883E-02 | 5.1640E-03 | 8.2032E-02 | 3.4354E-02 | 2.6498E-02 | 5.1640E-03 | 5.1640E-03 | 2.7286E-02 | 2.6986E-02 | 1.0865E-01 | 5.6189E-03 | 5.1640E-03 |
| | Std | 6.9549E-01 | 9.3618E-06 | 2.1932E+00 | 3.0921E-05 | 4.0078E-05 | 9.3618E-06 | 9.3618E-06 | 7.3920E-01 | 7.2193E-01 | 2.9081E+00 | 4.9402E-05 | 9.3618E-06 |
| F5 | Avg | 3.8864E-04 | 5.5716E-07 | 7.3568E-04 | 5.9363E-04 | 8.9553E-05 | 5.5716E-07 | 5.5716E-07 | 2.2349E-04 | 4.3761E-04 | 5.7315E-04 | 1.5798E-04 | 5.5716E-07 |
| | Std | 3.3269E-05 | 7.2768E-13 | 1.6135E-04 | 1.3124E-04 | 1.0231E-07 | 7.2768E-13 | 7.2768E-13 | 2.3907E-06 | 2.9959E-05 | 7.6346E-05 | 9.2823E-07 | 7.2768E-13 |
| F7 | Avg | 1.1157E-04 | 1.2076E-05 | 1.2361E-04 | 1.2897E-04 | 1.2638E-04 | 1.2076E-05 | 1.2076E-05 | 1.2369E-04 | 1.2223E-04 | 1.2228E-04 | 1.2828E-04 | 1.2076E-05 |
| | Std | 1.3064E-08 | 1.4328E-09 | 1.3922E-08 | 1.3623E-08 | 1.4328E-09 | 1.4328E-09 | 1.4328E-09 | 1.4890E-08 | 1.2846E-08 | 1.4201E-08 | 1.4353E-08 | 1.4328E-09 |
| F8 | Avg | -1.2231E+04 | -1.2527E+04 | -1.2211E+04 | -1.2140E+04 | -1.2166E+04 | -1.2527E+04 | -1.2527E+04 | -1.2234E+04 | -1.2284E+04 | -1.2184E+04 | -1.2233E+04 | -1.2527E+04 |
| | Std | 4.1838E+05 | 8.6230E+03 | 4.0985E+05 | 5.5995E+05 | 5.0951E+05 | 8.6230E+03 | 8.6230E+03 | 4.0300E+05 | 2.7943E+05 | 5.1486E+05 | 3.9926E+05 | 8.6230E+03 |
| F9 | Avg | 0.0000E+00 | 0.0000E+00 | 0.0000E+00 | 0.0000E+00 | 0.0000E+00 | 0.0000E+00 | 0.0000E+00 | 0.0000E+00 | 0.0000E+00 | 0.0000E+00 | 0.0000E+00 | 0.0000E+00 |
| | Std | 0.0000E+00 | 0.0000E+00 | 0.0000E+00 | 0.0000E+00 | 0.0000E+00 | 0.0000E+00 | 0.0000E+00 | 0.0000E+00 | 0.0000E+00 | 0.0000E+00 | 0.0000E+00 | 0.0000E+00 |
| F10 | Avg | 8.8818E-16 | 8.8818E-16 | 8.8818E-16 | 8.8818E-16 | 8.8818E-16 | 8.8818E-16 | 8.8818E-16 | 8.8818E-16 | 8.8818E-16 | 8.8818E-16 | 8.8818E-16 | 8.8818E-16 |
| | Std | 0.0000E+00 | 0.0000E+00 | 0.0000E+00 | 0.0000E+00 | 0.0000E+00 | 0.0000E+00 | 0.0000E+00 | 0.0000E+00 | 0.0000E+00 | 0.0000E+00 | 0.0000E+00 | 0.0000E+00 |
| F11 | Avg | 0.0000E+00 | 0.0000E+00 | 0.0000E+00 | 0.0000E+00 | 0.0000E+00 | 0.0000E+00 | 0.0000E+00 | 0.0000E+00 | 0.0000E+00 | 0.0000E+00 | 0.0000E+00 | 0.0000E+00 |
| | Std | 0.0000E+00 | 0.0000E+00 | 0.0000E+00 | 0.0000E+00 | 0.0000E+00 | 0.0000E+00 | 0.0000E+00 | 0.0000E+00 | 0.0000E+00 | 0.0000E+00 | 0.0000E+00 | 0.0000E+00 |
| F12 | Avg | 3.8504E-05 | 5.7613E-09 | 6.4453E-05 | 2.2367E-05 | 3.2174E-05 | 5.7613E-09 | 5.7613E-09 | 3.9925E-05 | 3.7442E-05 | 6.3313E-05 | 4.4636E-05 | 5.7613E-09 |
| | Std | 1.6973E-07 | 2.5338E-15 | 3.1122E-07 | 3.1268E-08 | 1.3129E-07 | 2.5338E-15 | 2.5338E-15 | 1.2126E-07 | 1.6751E-07 | 3.9589E-07 | 1.4654E-07 | 2.5338E-15 |
| F13 | Avg | 2.5268E-05 | 7.3758E-08 | 5.6365E-05 | 1.1599E-05 | 4.0506E-05 | 7.3758E-08 | 7.3758E-08 | 5.3111E-05 | 1.4429E-04 | 2.7472E-04 | 3.3519E-04 | 7.3758E-08 |
| | Std | 2.5417E-07 | 1.1564E-14 | 6.1552E-07 | 1.2537E-07 | 5.5799E-07 | 1.1564E-14 | 1.1564E-14 | 6.4852E-06 | 1.1973E-05 | 1.6274E-05 | 3.1603E-05 | 1.1564E-14 |
| F14 | Avg | 1.6271E+00 | 1.2568E+00 | 1.6300E+00 | 1.6292E+00 | 1.6331E+00 | 1.2568E+00 | 1.2568E+00 | 1.6647E+00 | 1.7016E+00 | 1.6729E+00 | 1.6351E+00 | 1.2568E+00 |
| | Std | 8.1281E-01 | 4.2014E-01 | 8.1418E-01 | 7.5186E-01 | 7.5475E-01 | 4.2014E-01 | 4.2014E-01 | 8.6754E-01 | 8.1706E-01 | 7.5072E-01 | 7.7219E-01 | 4.2014E-01 |
| F15 | Avg | 6.0018E-04 | 3.1114E-04 | 4.8056E-04 | 4.7329E-04 | 6.1174E-04 | 3.1114E-04 | 3.1114E-04 | 4.8231E-04 | 4.7794E-04 | 6.2905E-04 | 6.1847E-04 | 3.1114E-04 |
| | Std | 5.7123E-08 | 5.9227E-09 | 4.2229E-08 | 4.2130E-08 | 6.4023E-08 | 5.9227E-09 | 5.9227E-09 | 3.9255E-08 | 3.7585E-08 | 6.3280E-08 | 6.0363E-08 | 5.9227E-09 |
| F16 | Avg | -1.0316E+00 | -1.0316E+00 | -1.0316E+00 | -1.0316E+00 | -1.0316E+00 | -1.0316E+00 | -1.0316E+00 | -1.0316E+00 | -1.0316E+00 | -1.0316E+00 | -1.0316E+00 | -1.0316E+00 |
| | Std | 3.7659E-28 | 2.0454E-28 | 3.6468E-28 | 4.1299E-27 | 8.8193E-28 | 2.0454E-28 | 2.0454E-28 | 9.4225E-26 | 2.0568E-22 | 5.3015E-27 | 1.8193E-26 | 2.0454E-28 |
| F17 | Avg | 3.9789E-01 | 3.9789E-01 | 3.9789E-01 | 3.9789E-01 | 3.9789E-01 | 3.9789E-01 | 3.9789E-01 | 3.9789E-01 | 3.9789E-01 | 3.9789E-01 | 3.9789E-01 | 3.9789E-01 |
| | Std | 6.4879E-28 | 5.4939E-28 | 2.6873E-27 | 1.3948E-27 | 6.4956E-28 | 5.4939E-28 | 5.4939E-28 | 1.1243E-25 | 7.6632E-23 | 1.3581E-19 | 6.5868E-20 | 5.4939E-28 |
| F18 | Avg | 3.2166E+00 | 3.0270E+00 | 3.1626E+00 | 3.2703E+00 | 3.0815E+00 | 3.0270E+00 | 3.0270E+00 | 3.4059E+00 | 3.3792E+00 | 3.2974E+00 | 3.2169E+00 | 3.0270E+00 |
| | Std | 5.7851E+01 | 7.2827E-10 | 4.3476E+01 | 7.2169E+01 | 2.1804E+01 | 7.2827E-10 | 7.2827E-10 | 1.0770E+01 | 1.0062E+01 | 7.9305E-01 | 5.7851E+01 | 7.2827E-10 |
| F19 | Avg | -3.8628E+00 | -3.8628E+00 | -3.8628E+00 | -3.8628E+00 | -3.8628E+00 | -3.8628E+00 | -3.8628E+00 | -3.8628E+00 | -3.8628E+00 | -3.8628E+00 | -3.8628E+00 | -3.8628E+00 |
| | Std | 1.4141E-14 | 8.6647E-15 | 8.1016E-14 | 1.2650E-14 | 1.5864E-14 | 8.6647E-15 | 8.6647E-15 | 4.3868E-10 | 2.6346E-07 | 1.1324E-14 | 1.1343E-13 | 8.6647E-15 |
| F20 | Avg | -3.2751E+00 | -3.2713E+00 | -3.2749E+00 | -3.2772E+00 | -3.2726E+00 | -3.2713E+00 | -3.2713E+00 | -3.2745E+00 | -3.2773E+00 | -3.2766E+00 | -3.2751E+00 | -3.2713E+00 |
| | Std | 3.4463E-03 | 3.5391E-03 | 3.4545E-03 | 3.6469E-03 | 3.6189E-03 | 3.5391E-03 | 3.5391E-03 | 3.5159E-03 | 3.5241E-03 | 3.4363E-03 | 3.4358E-03 | 3.5391E-03 |
| F21 | Avg | -1.0153E+01 | -1.0153E+01 | -1.0153E+01 | -1.0153E+01 | -1.0153E+01 | -1.0153E+01 | -1.0153E+01 | -1.0153E+01 | -1.0153E+01 | -1.0153E+01 | -1.0153E+01 | -1.0153E+01 |
| | Std | 2.7364E-20 | 1.9212E-20 | 3.2065E-20 | 5.3411E-20 | 8.9653E-20 | 1.9212E-20 | 1.9212E-20 | 2.3969E-18 | 2.7226E-15 | 3.1640E-13 | 2.4321E-13 | 1.9212E-20 |
| F22 | Avg | -1.0403E+01 | -1.0402E+01 | -1.0403E+01 | -1.0403E+01 | -1.0403E+01 | -1.0402E+01 | -1.0402E+01 | -1.0403E+01 | -1.0403E+01 | -1.0403E+01 | -1.0403E+01 | -1.0402E+01 |
| | Std | 6.1829E-20 | 5.5748E-20 | 5.6164E-20 | 1.4156E-19 | 7.5138E-20 | 5.5748E-20 | 5.5748E-20 | 1.9467E-18 | 6.0416E-16 | 5.0390E-13 | 4.0150E-13 | 5.5748E-20 |
| F23 | Avg | -1.0536E+01 | -1.0536E+01 | -1.0536E+01 | -1.0536E+01 | -1.0536E+01 | -1.0536E+01 | -1.0536E+01 | -1.0536E+01 | -1.0536E+01 | -1.0536E+01 | -1.0536E+01 | -1.0536E+01 |
| | Std | 5.4597E-20 | 3.9712E-20 | 4.6431E-20 | 2.0897E-19 | 8.4799E-20 | 3.9712E-20 | 3.9712E-20 | 2.4598E-18 | 1.2705E-16 | 2.8091E-13 | 3.1066E-13 | 3.9712E-20 |

increases from 0.4 to 0.6, the values of $W$, $P_1$, and $P_3$ are set as 2.5, 0.6 and 0.6. When the value of $P_3$ increases from 0.4 to 0.6, the values of $W$, $P_1$, and $P_2$ are set as 2.5, 0.6 and 0.4. The effects of different parameters on the performance of HWEAVOA are shown in Table 2.

The results show that HWEAVOA has the highest search ability and algorithm stability when the values of $W$, $P_1$, $P_2$, and $P_3$ are set as 2.5, 0.6, 0.4 and 0.6, respectively. Therefore, the HWEAVOA selects the same parameter settings as the AVOA in this paper.

Based on the above-mentioned analysis, the main parameters of HWEAVOA and its variant algorithms are shown as Table 3.

Table 3. The parameter settings.

| Algorithm | Time/Year | Value |
|---|---|---|
| GA | 1980 | $P_c$=0.8, $P_m$=0.1 |
| PSO | 1995 | Inertia factor = 0.3, $c_1 = 1$, $c_2 = 1$ |
| DE | 1997 | Scaling factor = 0.5, Crossover probability = 0.5 |
| GWO | 2014 | $a = [2,0]$ |
| COOT | 2021 | / |
| RSO | 2021 | $R = floor\,((5-1) \times rand\,(1,1) + 1)$ |
| GTO | 2021 | p=0.03, $\beta = 3$, $\omega = 0.8$ |
| AOA | 2019 | $E_0 = [-1,1]$ |
| AVOA | 2021 | $p_1 = 0.6$, $p_2 = 0.4$, $p_3 = 0.6$, $\alpha = 0.8$, $\beta = 0.2$, $\gamma = 2.5$ |
| IHAOAVOA | 2022 | $p_1 = 0.6$, $p_2 = 0.4$, $p_3 = 0.6$, $\alpha = 0.8$, $\beta = 0.2$, $\gamma = 2.5$, $U = 0.00565$, $r_1 = 10$, $\omega = 0.005$, $\theta = 1.5\pi$ |
| OAVOA | 2022 | $L_1 = 0.8$, $L_2 = 0.2$, $w = 2.0$, $p_1 = 0.5$, $p_2 = 0.5$, $p_3 = 0.5$ |
| HAVOA | / | $p_1 = 0.6$, $p_2 = 0.4$, $p_3 = 0.6$, $\alpha = 0.8$, $\beta = 0.2$, $W = 2.5$, $a=1.4$, $b=0.3$ |
| WAVOA | / | $p_1 = 0.6$, $p_2 = 0.4$, $p_3 = 0.6$, $\alpha = 0.8$, $\beta = 0.2$, $W = 2.5$ |
| EAVOA | / | $p_1 = 0.6$, $p_2 = 0.4$, $p_3 = 0.6$, $\alpha = 0.8$, $\beta = 0.2$, $W = 2.5$ |
| HWAVOA | / | $p_1 = 0.6$, $p_2 = 0.4$, $p_3 = 0.6$, $\alpha = 0.8$, $\beta = 0.2$, $W = 2.5$, $a=1.4$, $b=0.3$ |
| HEAVOA | / | $p_1 = 0.6$, $p_2 = 0.4$, $p_3 = 0.6$, $\alpha = 0.8$, $\beta = 0.2$, $W = 2.5$, $a=1.4$, $b=0.3$ |
| WEAVOA | / | $p_1 = 0.6$, $p_2 = 0.4$, $p_3 = 0.6$, $\alpha = 0.8$, $\beta = 0.2$, $W = 2.5$ |
| HWEAVOA | / | $p_1 = 0.6$, $p_2 = 0.4$, $p_3 = 0.6$, $\alpha = 0.8$, $\beta = 0.2$, $W = 2.5$, $a=1.4$, $b=0.3$ |

**4.2 Effects of HCE, NWF and RLC**

To study the optimization performance of the HWEAVOA algorithm, three improved strategies (HCE, NWF and RLC) are introduced in the optimization process of HWEAVOA. In this section, we focus on the impact of the three improvement strategies on the original AVOA. The three improvement strategies are named briefly with "H", "W", and "E". The performance of HWEAVOA, 6 variants of AVOA: HAVOA, WAVOA, EAVOA, HWAVOA, HEAVOA, WEAVOA and AVOA is tested.

HAVOA, WAVOA and EAVOA represent the introduction of HCE, NWF and RLC into the original AVOA algorithm respectively. HWAVOA means the introduction of both HCE and NWF into the original AVOA algorithm. HEAVOA means to the introduction of both HCE and RLC into the original AVOA algorithm. WEAVOA means to the introduction of both NWF and RLC into the original AVOA algorithm. The parameter settings of these variant algorithms are shown in Table 3.

As shown in Table 4, the experimental results show that the optimal value of functions, F1~F4 can be obtained by HWEAVOA in unimodal benchmark functions. For multimodal benchmark functions, the optimal values of functions F8 and F11 can be obtained by HWEAVOA, while the optimal value found by HWEAVOA is the best of all algorithms on other multimodal benchmark functions. For fixed-dimension multimodal benchmark functions, the optimal values of functions F16, F17, F19, F21, F22 and F23 can be obtained by HWEAVOA, while HWEAVOA performs better on other fixed-dimensional multimodal benchmark functions than different algorithms.

As shown in Fig. 5, some convergence curves are displayed in 0~50 generations to show the difference in iterative efficiency more clearly and intuitively. When HWEAVOA solves F1 and F4, the convergence speed of the HWEAVOA algorithm is close to WAVOA, HWAVOA, and WEAVOA, which is significantly faster than HAVOA, EVOA. In the process of solving F6 ~ F8, F12, F13 and F15, the convergence speed of HWEAVOA is slightly faster than other variant algorithms. When solving F9, the optimal value is obtained around 30 times by all algorithms, but the convergence rate of HWEAVOA is still the fastest. When solving F21 and F23, the convergence speeds of HAVOA, EAVOA, HEAVOA, WEAVOA, AVOA and HWEAVOA are significantly faster than those of WAVOA and HWAVOA. Meanwhile, in HAVOA, EAVOA, HEAVOA, WEAVOA, AVOA and HWEAVOA, the convergence speed of HWEAVOA is slightly faster than those of HAVOA, EAVOA, HEAVOA, WEAVOA and AVOA. In summary, HWEAVOA has the advantage of the convergence speed and precision in the process of solving the classical test functions compared with other variants.

Table 4. The comparison results of HWEAVOA and its variant algorithms based on 23 classical functions.

| | | AVOA | HAVOA | WAVOA | EAVOA | HWAVOA | HEAVOA | WEAVOA | HWEAVOA |
|---|---|---|---|---|---|---|---|---|---|
| F1 | Avg | 1.9963E-153 | 2.5018E-148 | 0.0000E+00 | 4.0257E-158 | 0.0000E+00 | 2.5129E-148 | 0.0000E+00 | 0.0000E+00 |
| | Std | 2.4765E-303 | 6.2219E-293 | 0.0000E+00 | 3.8824E-313 | 0.0000E+00 | 6.2917E-293 | 0.0000E+00 | 0.0000E+00 |
| F2 | Avg | 3.8867E-82 | 8.4219E-82 | 0.0000E+00 | 3.9294E-78 | 0.0000E+00 | 7.8591E-82 | 0.0000E+00 | 0.0000E+00 |
| | Std | 1.2054E-160 | 4.6613E-160 | 0.0000E+00 | 1.5256E-152 | 0.0000E+00 | 5.9964E-160 | 0.0000E+00 | 0.0000E+00 |
| F3 | Avg | 8.9768E-102 | 5.4945E-110 | 0.0000E+00 | 2.1168E-107 | 0.0000E+00 | 7.6219E-114 | 0.0000E+00 | 0.0000E+00 |
| | Std | 8.0424E-200 | 8.6277E-217 | 0.0000E+00 | 4.4617E-211 | 0.0000E+00 | 5.7274E-224 | 0.0000E+00 | 0.0000E+00 |
| F4 | Avg | 2.2498E-79 | 7.4318E-74 | 0.0000E+00 | 1.2061E-77 | 0.0000E+00 | 1.3964E-75 | 0.0000E+00 | 0.0000E+00 |
| | Std | 2.9731E-155 | 5.5111E-144 | 0.0000E+00 | 1.0767E-151 | 0.0000E+00 | 9.5524E-148 | 0.0000E+00 | 0.0000E+00 |
| F5 | Avg | 1.0696E-01 | 1.8466E-01 | 9.5919E-01 | 8.0085E-02 | 8.4488E-01 | 5.3555E-02 | 2.7484E-02 | 5.1640E-03 |
| | Std | 2.8334E+00 | 4.9171E+00 | 2.5307E+01 | 2.1173E+00 | 2.2462E+01 | 1.4159E+00 | 7.1551E-01 | 9.3618E-06 |
| F6 | Avg | 9.4805E-04 | 6.9788E-04 | 4.8070E-03 | 4.2223E-06 | 3.6125E-03 | 4.3258E-06 | 9.9471E-04 | 5.5716E-07 |
| | Std | 2.1433E-04 | 1.4511E-04 | 7.8387E-04 | 2.1634E-11 | 5.6410E-04 | 1.6394E-11 | 1.7106E-04 | 7.2768E-13 |
| F7 | Avg | 2.6506E-04 | 2.7468E-04 | 1.2182E-04 | 2.6274E-04 | 1.1248E-04 | 2.8227E-04 | 1.1858E-04 | 1.2076E-05 |
| | Std | 8.6811E-08 | 8.2349E-08 | 1.5128E-08 | 7.0647E-08 | 1.2867E-08 | 8.8468E-08 | 1.4729E-08 | 1.4328E-09 |
| F8 | Avg | -1.2259E+04 | -1.2360E+04 | -1.1980E+04 | -1.1904E+04 | -1.2227E+04 | -1.2200E+04 | -1.1452E+04 | -1.2527E+04 |
| | Std | 3.3662E+05 | 82.4518E+05 | 7.3254E+05 | 7.6739E+05 | 3.6306E+05 | 4.9240E+05 | 1.2462E+06 | 8.6230E+03 |
| F9 | Avg | 0.0000E+00 | 0.0000E+00 | 0.0000E+00 | 0.0000E+00 | 0.0000E+00 | 0.0000E+00 | 0.0000E+00 | 0.0000E+00 |
| | Std | 0.0000E+00 | 0.0000E+00 | 0.0000E+00 | 0.0000E+00 | 0.0000E+00 | 0.0000E+00 | 0.0000E+00 | 0.0000E+00 |
| F10 | Avg | 8.8818E-16 | 8.8818E-16 | 8.8818E-16 | 8.8818E-16 | 8.8818E-16 | 8.8818E-16 | 8.8818E-16 | 8.8818E-16 |
| | Std | 0.0000E+00 | 0.0000E+00 | 0.0000E+00 | 0.0000E+00 | 0.0000E+00 | 0.0000E+00 | 0.0000E+00 | 0.0000E+00 |
| F11 | Avg | 0.0000E+00 | 0.0000E+00 | 0.0000E+00 | 0.0000E+00 | 0.0000E+00 | 0.0000E+00 | 0.0000E+00 | 0.0000E+00 |
| | Std | 0.0000E+00 | 0.0000E+00 | 0.0000E+00 | 0.0000E+00 | 0.0000E+00 | 0.0000E+00 | 0.0000E+00 | 0.0000E+00 |
| F12 | Avg | 4.9655E-05 | 2.8069E-05 | 2.0457E-04 | 1.8675E-07 | 2.7246E-04 | 1.9578E-07 | 5.3324E-05 | 5.7613E-09 |
| | Std | 3.6391E-07 | 1.3827E-07 | 1.4639E-06 | 2.3942E-14 | 2.2589E-06 | 3.3657E-14 | 2.2817E-07 | 2.5338E-15 |
| F13 | Avg | 1.1214E-04 | 2.5308E-04 | 7.4559E-04 | 1.2567E-05 | 8.2935E-04 | 1.2357E-05 | 4.4779E-04 | 7.3758E-08 |
| | Std | 1.0033E-05 | 2.0079E-05 | 1.0026E-04 | 1.2832E-07 | 3.3927E-04 | 1.4658E-07 | 2.5079E-05 | 1.1564E-14 |
| F14 | Avg | 1.6611E+00 | 1.6072E+00 | 1.9801E+00 | 1.4069E+00 | 1.9191E+00 | 1.3196E+00 | 1.6598E+00 | 1.2568E+00 |
| | Std | 2.6207E+00 | 2.2397E+00 | 4.4085E+00 | 5.4571E-01 | 4.1545E+00 | 4.4978E-01 | 8.0963E-01 | 4.2014E-01 |
| F15 | Avg | 4.8436E-04 | 4.9354E-04 | 5.9508E-04 | 4.7545E-04 | 5.9045E-04 | 4.8459E-04 | 6.0704E-04 | 3.1114E-04 |
| | Std | 4.3216E-08 | 4.6153E-08 | 4.8267E-08 | 4.4021E-08 | 4.9211E-08 | 4.4568E-08 | 5.7868E-08 | 5.9227E-09 |
| F16 | Avg | -1.0316E+00 | -1.0316E+00 | -1.0316E+00 | -1.0316E+00 | -1.0316E+00 | -1.0316E+00 | -1.0316E+00 | -1.0316E+00 |
| | Std | 2.0127E-27 | 1.3822E-27 | 3.5962E-19 | 6.4619E-28 | 9.3261E-20 | 2.4523E-27 | 4.5714E-19 | 2.0454E-28 |
| F17 | Avg | 3.9789E-01 | 3.9789E-01 | 3.9789E-01 | 3.9789E-01 | 3.9789E-01 | 3.9789E-01 | 3.9789E-01 | 3.9789E-01 |
| | Std | 1.1963E-27 | 6.1753E-28 | 2.9267E-19 | 8.8214E-28 | 2.3144E-19 | 9.8867E-28 | 1.1516E-19 | 5.4939E-28 |
| F18 | Avg | 3.0540E+00 | 3.0813E+00 | 3.6767E+00 | 3.0000E+00 | 3.2166E+00 | 3.0000E+00 | 3.1084E+00 | 3.0270E+00 |
| | Std | 1.4551E+00 | 2.1804E+00 | 2.6518E+01 | 7.0034E+00 | 5.7851E+00 | 4.9928E-10 | 2.9043E+00 | 7.2827E-10 |
| F19 | Avg | -3.8627E+00 | -3.8627E+00 | -3.8621E+00 | -3.8628E+00 | -3.8628E+00 | -3.8628E+00 | -3.8628E+00 | -3.8628E+00 |
| | Std | 3.3896E-07 | 2.8124E-15 | 4.0078E-06 | 2.3961E-14 | 3.2387E-06 | 4.5639E-15 | 1.5848E-14 | 8.6647E-15 |
| F20 | Avg | -3.2721E+00 | -3.2744E+00 | -3.2563E+00 | -3.2746E+00 | -3.2600E+00 | -3.2729E+00 | -3.2700E+00 | -3.2713E+00 |
| | Std | 3.7345E-03 | 3.5764E-03 | 8.0432E-03 | 3.3988E-03 | 1.0064E-02 | 3.3433E-03 | 3.5417E-03 | 3.5391E-03 |
| F21 | Avg | -1.0153E+01 | -1.0153E+01 | -1.0153E+01 | -1.0153E+01 | -1.0153E+01 | -1.0153E+01 | -1.0153E+01 | -1.0153E+01 |
| | Std | 4.7988E-20 | 2.0407E-20 | 2.8514E-01 | 2.8657E-20 | 2.7672E-01 | 4.2593E-19 | 6.0129E-13 | 1.9212E-20 |
| F22 | Avg | -1.0402E+01 | -1.0402E+01 | -1.0402E+01 | -1.0402E+01 | -1.0402E+01 | -1.0402E+01 | -1.0402E+01 | -1.0402E+01 |
| | Std | 2.4459E-20 | 5.5798E-20 | 5.5694E-01 | 4.0875E-20 | 3.8749E-01 | 1.6657E-20 | 7.1757E-13 | 5.5748E-20 |
| F23 | Avg | -1.0536E+01 | -1.0536E+01 | -1.0536E+01 | -1.0536E+01 | -1.0536E+01 | -1.0536E+01 | -1.0536E+01 | -1.0536E+01 |
| | Std | 5.0758E-20 | 5.2697E-20 | 4.3976E-01 | 6.7956E-20 | 2.9312E-01 | 5.6657E-20 | 5.3071E-13 | 3.9712E-20 |

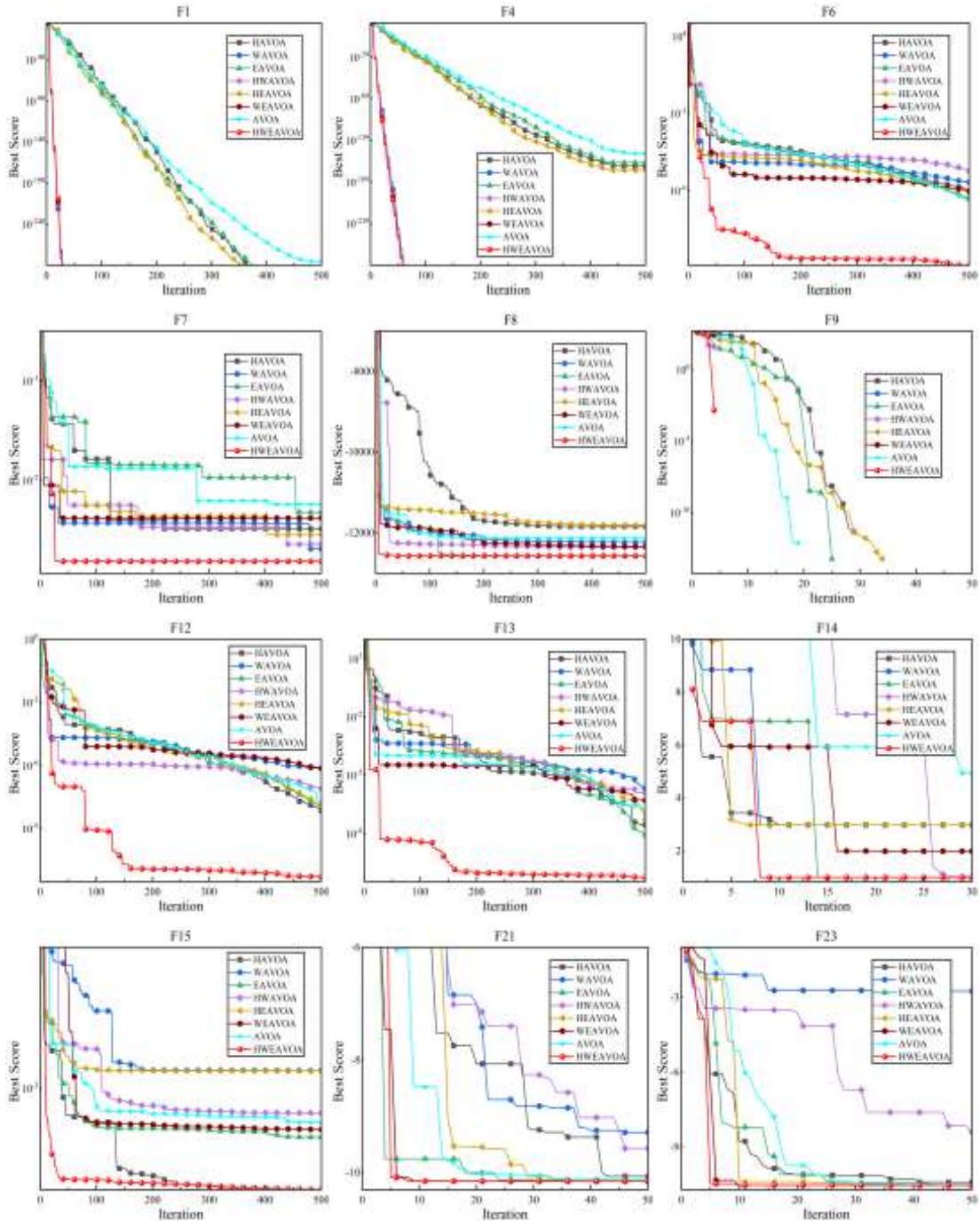

Fig. 5. Convergence curves of HWEAVOA and its variant algorithms based on F1, F4, F6, F7, F8, F9, F12, F13, F14, F15, F21 and F23.

**4.3 Evaluation with the classical test functions**

The performance of HWEAVOA is compared with 10 advanced intelligent optimization algorithms to verify the superiority, including GA (Deng, Zhang, et al., 2022), PSO (Cheng & Jin, 2015), DE (Das & Suganthan, 2011), GWO (Mirjalili, Mirjalili, & Lewis, 2014), COOT (Naruei & Keynia, 2021), RSO (Dhiman, Garg, Nagar, Kumar, & Dehghani, 2021), ATO (Naruei & Keynia, 2021), AOA (Abualigah, Diabat, Mirjalili, Abd Elaziz, & Gandomi, 2021), IHAOAVOA (Xiao, et al., 2022) and OAVOA (Jena, Naik, Panda, & Abraham, 2022). The parameter settings of the algorithms are shown in Table 3. The comparative results of these algorithms based on the 23 functions are shown in Table 5. For fairness, all used optimization methods run in the same test conditions.

The experimental results in Table 5 present that the HWEAVOA ranks first in 18 test functions of F1~F7, F9~F12, F15~F17, F19, and F21~F23, including seven unimodal benchmark functions, four multimodal benchmark functions and seven fixed-dimension multimodal benchmark functions. The performance of the HWEAVOA in classical test functions indicates its ability in solving real-world optimization problems. Especially for the IHAOAVOA and OAVOA, the HWEAVOA performs better than IHAOAVOA in 7 benchmark functions, including F5~F8, F12 F15 and F18, and better than OAVOA in F1~F7, F12, F13 and F15, which benefiting from the NWF strategy balances the exploration and exploitation capabilities, it is like a powerful engine that drives the HWEAVOA to excavate small areas effectively.

In order to have a more obvious analysis, the Friedman test is used to compare the solution results of the above algorithms. As shown in Table 6, the overall average ranking of HWEAVOA is 1.70, lower than the 1.96 of IHAOAVOA and the 2.43 of OAVOA. Meanwhile, it is evident that the HWEAVOA bestows a much faster convergence speed than other advanced algorithms as shown in Fig. 6. When solving F1, F3, F5, F7, F12 and F13, the HWEAVOA algorithm converges the fastest, followed by IHAOAVOA, and significantly faster than the other advanced algorithms, which means that it reaches to the best area very fast. It shows that the search performance, solution precision and the convergence velocity of the HWEAVOA are significantly enhanced. These features mean that the HWEAVOA has a robust global search capability, making an appropriate balance between the search mechanisms.

Table 5. The comparison results of HWEAVOA and other advanced algorithms based on 23 classical functions.

|  |  | GA | PSO | DE | GWO | COOT | RSO | GTO | AOA | IHAOAVOA | OAVOA | HWEAVOA |
|---|---|---|---|---|---|---|---|---|---|---|---|---|
| F1 | Avg | 4.1042E+04 | 1.2250E+01 | 3.6098E+04 | 4.7134E-15 | 2.7198E-08 | 8.6636E-248 | 6.0837E-220 | 4.8743E-10 | 0.0000E+00 | 1.3217E-214 | 0.0000E+00 |
|  | Std | 3.1080E+07 | 1.5256E+03 | 5.4347E+07 | 6.5329E-29 | 3.4832E-13 | 0.0000E+00 | 0.0000E+00 | 1.3742E-16 | 0.0000E+00 | 0.0000E+00 | 0.0000E+00 |
| F2 | Avg | 1.8851E+13 | 6.2036E+00 | 2.3112E+07 | 1.5543E-09 | 1.6678E-05 | 5.2416E-138 | 7.4023E-114 | 6.9967E-186 | 0.0000E+00 | 4.3159E-117 | 0.0000E+00 |
|  | Std | 1.6312E+28 | 4.2789E+00 | 2.9810E+16 | 9.4647E-19 | 8.4832E-08 | 8.2089E-273 | 1.3868E-224 | 0.0000E+00 | 0.0000E+00 | 1.8628E-230 | 0.0000E+00 |
| F3 | Avg | 1.1237E+05 | 7.0413E+03 | 6.9524E+04 | 6.8183E-02 | 1.7416E-06 | 2.7439E-89 | 9.5042E-208 | 9.7847E-03 | 0.0000E+00 | 1.4899E-189 | 0.0000E+00 |
|  | Std | 8.0911E+08 | 2.7818E+07 | 2.4865E+08 | 3.0893E-02 | 3.0127E-09 | 7.5145E-175 | 0.0000E+00 | 3.9430E+04 | 0.0000E+00 | 0.0000E+00 | 0.0000E+00 |
| F4 | Avg | 6.7858E+01 | 3.3038E+00 | 8.6647E+01 | 9.4083E-04 | 1.2834E-04 | 2.4997E-35 | 1.0824E-113 | 3.0869E-02 | 0.0000E+00 | 5.8533E-106 | 0.0000E+00 |
|  | Std | 1.8298E+01 | 1.2591E+00 | 1.8375E+01 | 6.0127E-07 | 6.2719E-06 | 6.2148E-67 | 1.4027E-224 | 3.5147E-04 | 0.0000E+00 | 3.3842E-208 | 0.0000E+00 |
| F5 | Avg | 1.4811E+08 | 2.2561E+02 | 1.0701E+08 | 2.7391E+01 | 6.6792E+01 | 2.8826E+01 | 4.5404E+00 | 2.8567E+01 | 2.1256E-01 | 5.4877E-02 | 5.1640E-03 |
|  | Std | 9.6595E+14 | 1.8096E+04 | 1.8578E+15 | 6.0839E-01 | 9.0176E+03 | 3.9817E-02 | 9.4188E+01 | 6.5634E-02 | 5.6026E+00 | 1.4621E+00 | 9.3618E-06 |
| F6 | Avg | 4.1204E+04 | 1.1588E+01 | 3.6279E+04 | 1.0079E+01 | 1.1245E+00 | 3.4098E+00 | 7.1435E-05 | 3.5097E+00 | 3.9115E-06 | 3.7343E-05 | 5.5716E-07 |
|  | Std | 2.9506E+07 | 9.3303E+02 | 5.5716E+07 | 1.8281E-01 | 1.0970E+00 | 2.4009E-01 | 3.9127E-08 | 8.1452E-02 | 5.1962E-11 | 2.0141E-06 | 7.2768E-13 |
| F7 | Avg | 7.5735E+01 | 7.2691E-01 | 4.8078E+01 | 3.7954E-03 | 8.6162E-03 | 4.7203E-04 | 1.6827E-04 | 1.1423E-04 | 6.0195E-05 | 3.3083E-04 | 1.2076E-05 |
|  | Std | 2.4761E+02 | 9.3997E-02 | 3.6733E+02 | 3.9015E-06 | 5.1927E-05 | 3.3459E-07 | 1.9676E-08 | 1.2764E-08 | 3.8296E-09 | 1.2611E-07 | 1.4328E-09 |
| F8 | Avg | -5.3712E+03 | -2.5205E+03 | -3.0526E+03 | -5.9347E+03 | -6.9383E+03 | -5.6900E+03 | -1.2569E+04 | -5.0043E+03 | -1.2033E+04 | -1.2532E+04 | -1.2527E+04 |
|  | Std | 1.0771E+07 | 1.5915E+05 | 1.6331E+05 | 8.2120E+05 | 9.0514E+05 | 1.1531E+06 | 2.4892E-04 | 1.8333E+05 | 1.2818E+06 | 1.1926E+04 | 8.6230E+03 |
| F9 | Avg | 2.9866E+02 | 6.5850E+01 | 3.6020E+02 | 7.9611E+00 | 2.0334E-05 | 1.3200E-01 | 0.0000E+00 | 0.0000E+00 | 0.0000E+00 | 0.0000E+00 | 0.0000E+00 |
|  | Std | 5.7275E+02 | 2.3507E+02 | 7.1989E+02 | 4.9283E+01 | 1.2907E-07 | 1.7406E+01 | 0.0000E+00 | 0.0000E+00 | 0.0000E+00 | 0.0000E+00 | 0.0000E+00 |
| F10 | Avg | 1.7983E+01 | 5.7875E+00 | 1.9674E+01 | 1.2613E-08 | 4.4427E-06 | 1.9959E-02 | 8.8846E-16 | 8.8846E-16 | 8.8818E-16 | 8.8818E-16 | 8.8818E-16 |
|  | Std | 3.6111E-01 | 1.1971E+00 | 2.4807E-01 | 5.9814E-17 | 1.2249E-08 | 3.9797E-01 | 0.0000E+00 | 0.0000E+00 | 0.0000E+00 | 0.0000E+00 | 0.0000E+00 |
| F11 | Avg | 3.6948E+02 | 3.7399E+02 | 3.2642E+02 | 7.0424E-03 | 3.1865E-07 | 0.0000E+00 | 0.0000E+00 | 3.0226E-01 | 0.0000E+00 | 0.0000E+00 | 0.0000E+00 |
|  | Std | 2.4912E+03 | 1.1165E+03 | 4.8491E+03 | 1.4327E-04 | 7.0743E-11 | 0.0000E+00 | 0.0000E+00 | 3.7848E-02 | 0.0000E+00 | 0.0000E+00 | 0.0000E+00 |
| F12 | Avg | 3.4562E+08 | 2.2401E+00 | 2.3115E+08 | 6.0822E-02 | 2.6490E-01 | 3.5152E-01 | 3.4023E-06 | 5.9888E-01 | 4.2621E-08 | 1.6857E-07 | 5.7613E-09 |
|  | Std | 6.6870E+15 | 9.6440E-01 | 1.6850E+16 | 1.5193E-03 | 2.6387E-01 | 1.7652E-02 | 5.1433E-11 | 2.3017E-03 | 5.0366E-15 | 7.8192E-14 | 2.5338E-15 |
| F13 | Avg | 6.7274E+08 | 1.7115E+01 | 4.5862E+08 | 8.2798E-01 | 9.3609E-01 | 2.8840E+00 | 3.0871E-03 | 2.8447E+00 | 7.3259E-08 | 3.0287E-06 | 7.3758E-08 |
| | Std | 2.1793E+16 | 5.0229E+01 | 4.3448E+16 | 7.1988E-02 | 3.6735E-01 | 1.7200E-02 | 1.0143E-04 | 9.1592E-03 | 2.1563E-14 | 2.3512E-11 | 1.1564E-14 |
| F14 | Avg | 3.5323E+00 | 1.5661E+00 | 6.0050E+00 | 4.7868E+00 | 1.1190E+00 | 2.8793E+00 | 9.9800E-01 | 9.8511E+00 | 1.1240E+00 | 1.1450E+00 | 1.2568E+00 |
| | Std | 5.2759E+00 | 1.6942E+00 | 1.5198E+01 | 1.7402E+01 | 5.4012E-01 | 5.3491E+00 | 3.1427E-14 | 1.5635E+01 | 2.2331E-01 | 1.6562E-01 | 4.2014E-01 |
| F15 | Avg | 6.9147E-02 | 1.4523E-03 | 1.1645E-02 | 4.8843E-03 | 1.4215E-03 | 1.2709E-03 | 4.3286E-04 | 1.9243E-02 | 3.2956E-04 | 3.4519E-04 | 3.1114E-04 |
| | Std | 5.7091E-03 | 2.0914E-05 | 6.9245E-05 | 6.8234E-05 | 1.3566E-05 | 4.1096E-06 | 9.8458E-08 | 8.0452E-04 | 5.1782E-09 | 1.3427E-08 | 5.9227E-09 |
| F16 | Avg | -1.0316E+00 | -1.0316E+00 | -1.0073E+00 | -1.0316E+00 | -1.0316E+00 | -1.0314E+00 | -1.0316E+00 | -1.0316E+00 | -1.0316E+00 | -1.0316E+00 | -1.0316E+00 |
| | Std | 2.0421E-20 | 4.9449E-02 | 8.2801E-04 | 8.9123E-32 | 5.4623E-15 | 1.0546E-06 | 1.4023E-08 | 1.0924E-31 | 2.5641E-31 | 2.4987E-18 | 2.0454E-28 |
| F17 | Avg | 4.0333E-01 | 3.9923E-01 | 5.5624E-01 | 3.9789E-01 | 3.9789E-01 | 3.9789E-01 | 3.9789E-01 | 3.9789E-01 | 3.9789E-01 | 3.9789E-01 | 3.9789E-01 |
| | Std | 3.8374E-12 | 4.5679E-03 | 4.7895E-02 | 1.2345E-05 | 0.0000E+00 | 0.0000E+00 | 0.0000E+00 | 0.0000E+00 | 3.2654E-08 | 7.3978E-08 | 5.4939E-28 |
| F18 | Avg | 1.5813E+01 | 3.5473E+00 | 3.7994E+00 | 3.6481E+00 | 3.0000E+00 | 3.0001E+00 | 3.0000E+00 | 1.3381E+01 | 3.7290E+00 | 3.0000E+00 | 3.0270E+00 |
| | Std | 1.4722E+02 | 2.7410E+01 | 3.0565E+00 | 5.2068E+01 | 6.9277E-19 | 2.3944E-08 | 2.1145E-30 | 2.8020E+02 | 1.9151E+01 | 4.6328E-14 | 7.2827E-10 |
| F19 | Avg | -3.8616E+00 | -3.8609E+00 | -3.8465E+00 | -3.8614E+00 | -3.8627E+00 | -3.4179E+00 | -3.8627E+00 | -3.8497E+00 | -3.8628E+00 | -3.8628E+00 | -3.8628E+00 |
| | Std | 1.9023E-06 | 8.3956E-04 | 4.2195E-04 | 5.1753E-06 | 4.5023E-19 | 1.3269E-01 | 1.4459E-30 | 3.3421E-05 | 4.7624E-18 | 8.5026E-12 | 8.6647E-15 |
| F20 | Avg | -3.2653E+00 | -3.2493E+00 | -2.8880E+00 | -3.2590E+00 | -3.2921E+00 | -1.7429E+00 | -3.2763E+00 | -3.0254E+00 | -3.2731E+00 | -3.2813E+00 | -3.2713E+00 |
| | Std | 3.4172E-03 | 1.1971E-02 | 3.4913E-02 | 6.8934E-03 | 2.6569E-03 | 2.9658E-01 | 3.3445E-03 | 1.2822E-02 | 3.4848E-03 | 3.1829E-03 | 3.5391E-03 |
| F21 | Avg | -5.7021E+00 | -5.6647E+00 | -2.1978E+00 | -8.9422E+00 | -8.4477E+00 | -7.7425E-01 | -1.0153E+01 | -3.7305E+00 | -1.0153E+01 | -1.0153E+01 | -1.0153E+01 |
| | Std | 1.0379E+01 | 1.0981E+01 | 1.1273E+00 | 5.9845E+00 | 8.8103E+00 | 3.5984E-01 | 2.6654E-30 | 1.8135E+00 | 1.5897E-20 | 9.2574E-14 | 1.9212E-20 |
| F22 | Avg | -5.8522E+00 | -5.9267E+00 | -2.4080E+00 | -1.0209E+01 | -9.5557E+00 | -1.0310E+01 | -1.4029E+01 | -3.7478E+00 | -1.4029E+01 | -1.4029E+01 | -1.4029E+01 |
| | Std | 1.1314E+01 | 1.1321E+01 | 1.0497E+00 | 1.1966E+00 | 5.1247E+00 | 5.0322E-01 | 3.7216E-30 | 2.2685E+00 | 3.5447E-20 | 2.1959E-12 | 5.5748E-20 |
| F23 | Avg | -5.8181E+00 | -5.5764E+00 | -2.5292E+00 | -1.0226E+01 | -9.9543E+00 | -1.2656E+01 | -1.0536E+01 | -3.8141E+00 | -1.0536E+01 | -1.0536E+01 | -1.0536E+01 |
| | Std | 1.1950E+01 | 1.2440E+01 | 1.3680E+00 | 2.2807E+00 | 3.8712E+00 | 5.9897E-01 | 7.0713E-30 | 2.8314E+00 | 4.2365E-13 | 9.9654E-20 | 3.9712E-20 |

81

Table 6. The Friedman test results based on the classical test functions.

| | GA | PSO | DE | GWO | COOT | RSO | GTO | AOA | IHAOAVOA | OAVOA | HWEAVOA |
|---|---|---|---|---|---|---|---|---|---|---|---|
| F1 | 11 | 9 | 10 | 6 | 8 | 3 | 4 | 7 | 1 | 5 | 1 |
| F2 | 11 | 9 | 10 | 7 | 8 | 4 | 6 | 3 | 1 | 5 | 1 |
| F3 | 11 | 9 | 10 | 8 | 6 | 5 | 3 | 7 | 1 | 4 | 1 |
| F4 | 10 | 9 | 11 | 7 | 6 | 5 | 8 | 3 | 1 | 4 | 1 |
| F5 | 11 | 9 | 10 | 5 | 8 | 7 | 4 | 6 | 3 | 2 | 1 |
| F6 | 11 | 9 | 10 | 5 | 6 | 7 | 4 | 8 | 2 | 3 | 1 |
| F7 | 11 | 9 | 10 | 7 | 8 | 6 | 4 | 3 | 2 | 5 | 1 |
| F8 | 8 | 11 | 10 | 6 | 5 | 7 | 1 | 9 | 4 | 2 | 3 |
| F9 | 10 | 9 | 11 | 8 | 7 | 6 | 1 | 1 | 1 | 1 | 1 |
| F10 | 10 | 9 | 11 | 6 | 7 | 8 | 1 | 1 | 1 | 1 | 1 |
| F11 | 10 | 11 | 9 | 7 | 6 | 1 | 1 | 8 | 1 | 1 | 1 |
| F12 | 11 | 9 | 10 | 5 | 6 | 7 | 4 | 8 | 2 | 3 | 1 |
| F13 | 11 | 9 | 10 | 5 | 6 | 8 | 4 | 7 | 1 | 3 | 2 |
| F14 | 8 | 6 | 10 | 9 | 2 | 7 | 1 | 11 | 3 | 4 | 5 |
| F15 | 11 | 7 | 9 | 8 | 6 | 5 | 4 | 10 | 2 | 3 | 1 |
| F16 | 1 | 1 | 11 | 1 | 1 | 10 | 1 | 1 | 1 | 1 | 1 |
| F17 | 10 | 9 | 11 | 1 | 1 | 1 | 1 | 1 | 1 | 1 | 1 |
| F18 | 11 | 6 | 9 | 7 | 1 | 4 | 1 | 10 | 8 | 1 | 5 |
| F19 | 6 | 8 | 10 | 7 | 4 | 11 | 4 | 9 | 1 | 1 | 1 |
| F20 | 7 | 9 | 2 | 8 | 1 | 11 | 4 | 10 | 5 | 3 | 6 |
| F21 | 7 | 8 | 10 | 5 | 6 | 11 | 1 | 9 | 1 | 1 | 1 |
| F22 | 9 | 8 | 11 | 6 | 7 | 5 | 1 | 10 | 1 | 1 | 1 |
| F23 | 8 | 9 | 11 | 6 | 7 | 5 | 1 | 10 | 1 | 1 | 1 |
| Avg. Rank/No. | 9.30/10 | 8.35/9 | 9.83/11 | 6.09/6 | 5.35/5 | 6.26/7 | 2.78/4 | 6.61/8 | 1.96/2 | 2.43/3 | 1.70/1 |

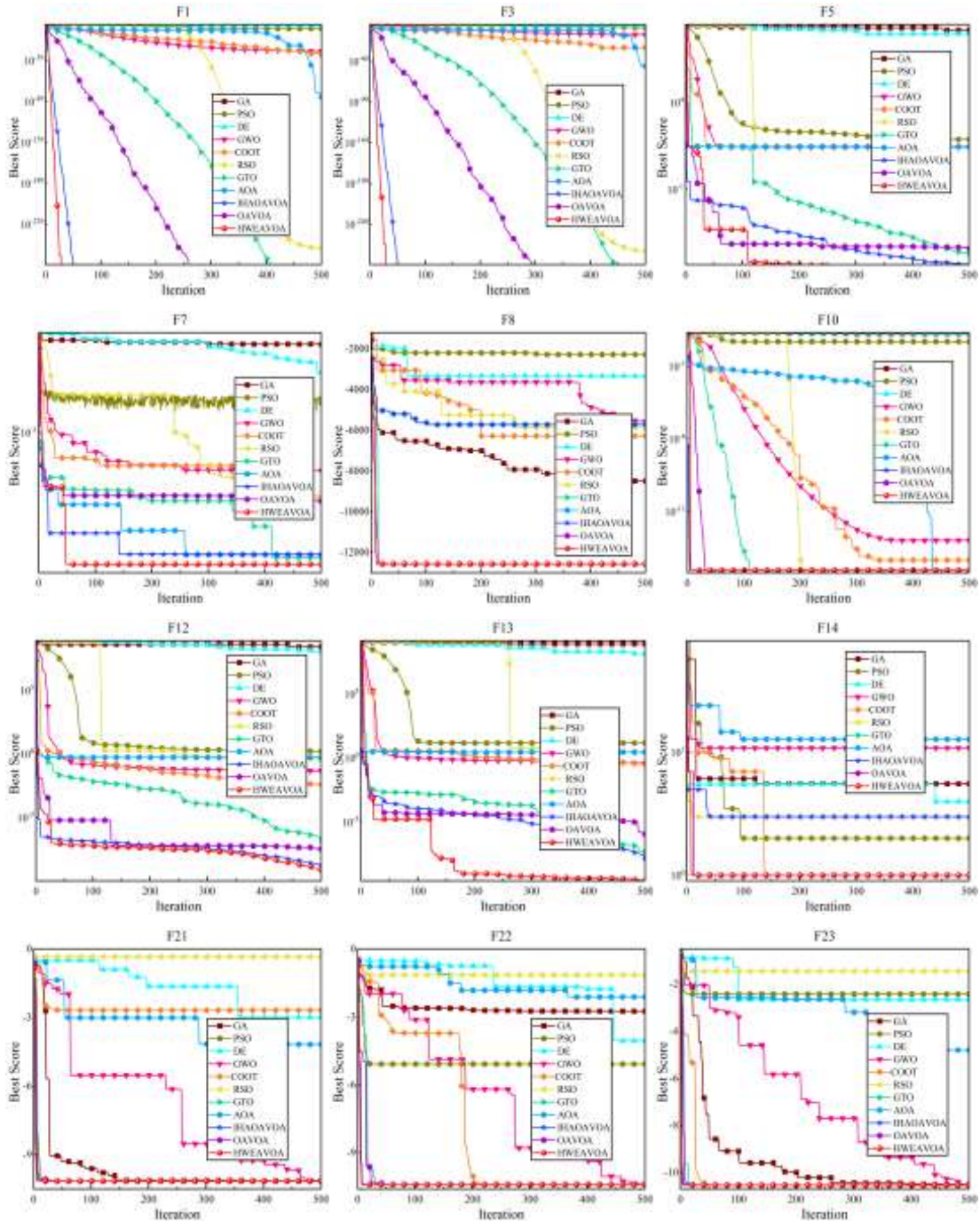

Fig. 6. Convergence curves of HWEAVOA and advanced algorithms based on F1, F3, F5, F7, F8, F10, F12, F13, F14, F21, F22 and F23.

**4.4 Evaluation with the CEC2022 test functions**

After using classical test functions to comprehensively evaluate the proposed HWEAVOA, the CEC2022 test functions (Dimension = 10 and 20) is applied to comprehensively evaluate the algorithm. The purpose of this section is to provide a reference for subsequent research, so that other scholars can compare and evaluate the AVOA based on the CEC2022 test functions. To be fair, all algorithms are performed under the same test conditions. The parameter settings of the algorithms are shown in Table 3.

In the CEC2022 test functions, F1 is the unimodal function, F2~F5 are basic functions, F6~F8 are hybrid functions, and F9~F12 are the composition functions. Different types of test functions can be used to evaluate the algorithm performance from different aspects. The experimental results of the HWEAVOA with 10 and 20 dimensions are shown in Table 7 and Table 8.

The experimental results in Table 7 and Table 8 present that the HWEAVOA can obtain the competitive results consistently in almost all test functions for both 10 and 20 dimensions successfully. For the whole CEC2022 test functions, the HWEAVOA's advantages are obvious, which is due that the HCE strategy makes the initial population distribution more homogeneous, and enhances the global optimization performance and convergence rate of the AVOA. At the same time, the RLC strategy increases the diversity of the population and chance of obtaining better solutions, the bad performed individuals are given learning opportunities to become dominant individuals.

To make the analysis more comprehensive, the Friedman test is used to compare the solution results of these advanced algorithms. As shown in Table 9, the overall average ranking of HWEAVOA in 10 dimension is 2.25, lower than the 2.83 of COOT and the 3.00 of OAVOA. Meanwhile, the overall average ranking of HWEAVOA in 10 dimension is 2.50, far lower than the 3.75 of COOT and the 4.17 of GTO. In summary, the HWEAVOA overcomes the shortcomings of the original AVOA and achieves better algorithm performance with the assistance of the HCE and RLC strategy.

**4.5 Population size analysis**

In this section, the influence of the population size is examined on the classical test functions. In order to adequately analyze the population size sensitivity of the HWEAVOA, the population size is set respectively to 30, 60, 100 and 200 to demonstrate its influence on the HWEAVOA.

The experimental data are shown in Table 10, where the Avg and Std values of HWEAVOA varied slightly from population size to population in 23 test functions, but the overall order of magnitude remained basically at the same level. Because there is little difference between the given population size, the claim as mentioned earlier is supported. Therefore, the HWEAVOA is more stable when the population changes.

Table 7. The comparison results based on the CEC2022 test functions in 10 dimensions.

| | | GA | PSO | DE | GWO | COOT | RSO | GTO | AOA | IHAOAVOA | OAVOA | HWEAVOA |
|---|---|---|---|---|---|---|---|---|---|---|---|---|
| F1 | Avg | 1.9995E+04 | 3.0001E+02 | -5.5927E+01 | 1.7534E+03 | 3.0000E+02 | 2.3436E+03 | 3.0000E+02 | 3.0001E+02 | 3.0000E+02 | 3.0000E+02 | 3.0000E+02 |
| | Std | 3.0080E+07 | 1.1900E-04 | 6.7143E-01 | 4.4890E+06 | 4.5500E-22 | 2.4512E+06 | 1.6200E-24 | 6.2700E-06 | 1.7200E-27 | 6.5719E-27 | 1.4100E-26 |
| F2 | Avg | 6.8903E+02 | 4.2197E+02 | 5.0036E+01 | 4.1862E+02 | 4.0380E+02 | 5.7541E+02 | 4.0562E+02 | 4.1036E+02 | 4.0722E+02 | 4.1151E+02 | 4.0448E+02 |
| | Std | 1.1520E+04 | 9.8515E+02 | 7.8571E+02 | 3.0141E+02 | 1.5297E+01 | 2.2229E+04 | 1.1312E+01 | 5.6712E+02 | 1.5625E+02 | 4.0075E+02 | 1.3888E+01 |
| F3 | Avg | 6.0021E+02 | 6.3422E+02 | 7.4124E+01 | 6.0970E+02 | 6.0474E+02 | 6.5138E+02 | 6.0636E+02 | 6.3708E+02 | 6.0737E+02 | 6.0138E+02 | 6.0174E+02 |
| | Std | 5.2720E-03 | 9.9069E+01 | 6.6654E+02 | 3.7110E+00 | 7.8997E+01 | 4.9371E+01 | 3.5148E+01 | 7.1256E+01 | 4.1175E+00 | 7.2203E-02 | 4.1847E+00 |
| F4 | Avg | 8.3484E+02 | 8.4777E+02 | 3.4137E+01 | 8.1298E+02 | 8.3482E+02 | 8.3254E+02 | 8.3383E+02 | 8.2769E+02 | 8.3383E+02 | 8.3482E+02 | 8.2537E+02 |
| | Std | 5.3886E+01 | 1.1947E+02 | 2.8070E+02 | 2.8330E+01 | 4.8599E+01 | 2.2008E+01 | 4.7252E+01 | 7.5699E+01 | 2.7978E+01 | 6.2494E+01 | 7.6737E+01 |
| F5 | Avg | 1.3542E+03 | 1.2216E+03 | -2.5214E+01 | 9.5237E+02 | 9.0002E+02 | 1.1271E+03 | 9.7532E+02 | 1.2902E+03 | 9.4161E+02 | 9.0023E+02 | 9.0099E+02 |
| | Std | 3.1483E+04 | 2.6976E+04 | 2.0908E+00 | 1.1218E+02 | 6.8190E-03 | 1.4193E+04 | 9.0682E+03 | 1.0642E+04 | 1.1198E+04 | 5.1482E-02 | 2.9881E+00 |
| F6 | Avg | 1.8176E+07 | 2.1505E+03 | 3.7966E+00 | 6.0399E+03 | 3.9331E+03 | 5.7324E+03 | 1.8094E+03 | 3.5362E+03 | 3.3599E+03 | 3.3628E+03 | 2.8883E+03 |
| | Std | 1.2712E+14 | 1.5391E+06 | 2.9026E+03 | 5.4581E+06 | 2.6374E+06 | 2.9482E+14 | 1.2809E+02 | 2.1892E+06 | 2.0490E+06 | 3.1717E+06 | 1.0477E+06 |
| F7 | Avg | 2.0208E+03 | 2.0636E+03 | 1.7459E+01 | 2.0241E+03 | 2.0181E+03 | 2.0634E+03 | 2.0267E+03 | 2.0829E+03 | 2.0190E+03 | 2.0199E+03 | 2.0138E+03 |
| | Std | 8.3430E+01 | 5.3316E+02 | 9.3407E+03 | 1.5463E+02 | 4.7923E+01 | 4.2172E+02 | 1.0953E+02 | 4.0626E+02 | 3.8705E+01 | 2.4507E+01 | 8.6743E+01 |
| F8 | Avg | 2.2392E+03 | 2.2698E+03 | -1.1462E+01 | 2.2208E+03 | 2.2221E+03 | 2.2370E+03 | 2.2209E+03 | 2.2908E+03 | 2.2168E+03 | 2.2162E+03 | 2.2179E+03 |
| | Std | 2.3366E+01 | 3.9262E+03 | 4.4070E+01 | 4.5022E+01 | 6.3696E+01 | 6.3623E+01 | 1.2209E+01 | 7.3968E+03 | 6.1120E+01 | 6.4969E+01 | 4.6032E+01 |
| F9 | Avg | 2.5427E+03 | 2.5401E+03 | -1.0000E+02 | 2.5531E+03 | 2.5293E+03 | 2.6010E+03 | 2.5293E+03 | 2.5337E+03 | 2.5293E+03 | 2.5293E+03 | 2.5293E+03 |
| | Std | 5.9104E+01 | 1.3256E+03 | 8.2500E-26 | 8.0224E+02 | 0.0000E+00 | 2.9113E+03 | 0.0000E+00 | 5.5817E+02 | 0.0000E+00 | 4.8344E-26 | 0.0000E+00 |
| F10 | Avg | 2.5008E+03 | 2.6670E+03 | -3.4148E+01 | 2.5747E+03 | 2.5003E+03 | 2.5084E+03 | 2.5262E+03 | 2.6651E+03 | 2.5003E+03 | 2.5003E+03 | 2.5003E+03 |
| | Std | 4.2841E-02 | 5.6401E+04 | 5.4604E+02 | 2.7634E+03 | 3.9770E-03 | 1.0054E+02 | 2.6568E+03 | 1.5394E+04 | 4.4440E-03 | 9.0221E-03 | 8.3947E-03 |
| F11 | Avg | 2.8061E+03 | 2.7291E+03 | 5.4929E+01 | 2.8138E+03 | 2.6000E+03 | 3.0906E+03 | 2.7505E+03 | 2.7508E+03 | 2.6500E+03 | 2.6267E+03 | 2.6684E+03 |
| | Std | 3.2922E+03 | 2.2252E+04 | 2.9914E+02 | 2.8021E+04 | 2.7700E-13 | 1.3164E+05 | 2.6160E+03 | 2.3527E+04 | 1.5006E+04 | 9.9556E+03 | 1.8921E+04 |
| F12 | Avg | 2.8706E+03 | 2.9848E+03 | -6.6854E+01 | 2.8878E+03 | 2.8612E+03 | 2.9397E+03 | 2.8684E+03 | 2.9947E+03 | 2.8666E+03 | 2.8659E+03 | 2.8644E+03 |
| | Std | 8.0265E+00 | 6.0686E+03 | 1.1489E+03 | 3.1006E+01 | 5.1857E+00 | 4.2442E+02 | 2.4881E+00 | 4.3692E+03 | 3.4936E+00 | 2.9429E+00 | 3.6782E+01 |

Table 8. The comparison results based on the CEC2022 test functions in 20 dimensions.

|  |  | GA | PSO | DE | GWO | COOT | RSO | GTO | AOA | IHAOAVOA | OAVOA | HWEAVOA |
|---|---|---|---|---|---|---|---|---|---|---|---|---|
| F1 | Avg | 7.6011E+04 | 7.0912E+02 | 8.1813E+02 | 6.6808E+03 | 3.0000E+02 | 1.4181E+04 | 3.0000E+02 | 3.0010E+02 | 3.0000E+02 | 3.0000E+02 | 3.0000E+02 |
|  | Std | 2.0700E+08 | 1.5229E+06 | 2.6637E+06 | 1.3043E+07 | 3.4400E-21 | 3.3025E+07 | 1.1300E-20 | 5.0100E-04 | 5.3900E-28 | 9.8600E-26 | 1.4000E-27 |
| F2 | Avg | 1.2841E+03 | 4.5225E+02 | 4.3727E+02 | 4.9385E+02 | 4.3273E+02 | 8.6792E+02 | 4.2849E+02 | 4.5527E+02 | 4.2497E+02 | 4.2916E+02 | 4.2891E+02 |
|  | Std | 3.9865E+04 | 6.4405E+02 | 6.0584E+02 | 1.2853E+03 | 5.2264E+02 | 5.3667E+04 | 5.4241E+02 | 1.1691E+02 | 5.5722E+02 | 5.4359E+02 | 5.8716E+02 |
| F3 | Avg | 6.0015E+02 | 6.5143E+02 | 6.0119E+02 | 6.0267E+02 | 6.0564E+02 | 6.6415E+02 | 6.3209E+02 | 6.5379E+02 | 6.0194E+02 | 6.0070E+02 | 6.0093E+02 |
|  | Std | 2.9210E-03 | 4.3689E+01 | 1.2426E+00 | 6.8511E+00 | 2.4144E+01 | 5.5494E+01 | 1.4569E+02 | 4.2390E+01 | 6.9376E+00 | 1.1823E+00 | 9.3945E-01 |
| F4 | Avg | 8.7881E+02 | 8.8546E+02 | 8.3256E+02 | 8.4287E+02 | 8.6192E+02 | 9.1696E+02 | 8.7668E+02 | 8.8276E+02 | 8.8832E+02 | 8.8245E+02 | 8.6169E+02 |
|  | Std | 7.9420E+02 | 4.7240E+02 | 2.2830E+02 | 1.3756E+02 | 2.8356E+02 | 2.1794E+02 | 2.7295E+02 | 3.3106E+02 | 1.6662E+02 | 2.1975E+02 | 1.6810E+02 |
| F5 | Avg | 2.9479E+03 | 2.2782E+03 | 9.6690E+02 | 1.1245E+03 | 1.1783E+03 | 2.3423E+03 | 1.8932E+03 | 2.3299E+03 | 2.3803E+03 | 2.2113E+03 | 2.1837E+03 |
|  | Std | 5.8630E+05 | 2.8727E+05 | 8.5202E+03 | 3.2871E+04 | 4.2179E+04 | 6.2006E+04 | 1.8306E+05 | 7.2733E+04 | 2.7854E+04 | 8.0193E+04 | 7.9051E+04 |
| F6 | Avg | 2.9100E+08 | 2.7457E+03 | 8.9529E+03 | 3.5561E+04 | 5.1210E+03 | 7.2939E+07 | 2.1717E+03 | 5.1852E+03 | 3.7880E+03 | 5.3965E+03 | 3.9485E+03 |
|  | Std | 1.2400E+16 | 1.3924E+06 | 4.7130E+07 | 7.2722E+09 | 1.2043E+07 | 6.6014E+15 | 2.1407E+05 | 4.3895E+06 | 6.7910E+06 | 2.5995E+07 | 9.7389E+06 |
| F7 | Avg | 2.0340E+03 | 2.1553E+03 | 2.0493E+03 | 2.0499E+03 | 2.0497E+03 | 2.1451E+03 | 2.0888E+03 | 2.2145E+03 | 2.0514E+03 | 2.0496E+03 | 2.0483E+03 |
|  | Std | 1.4750E+02 | 2.4396E+03 | 9.1386E+02 | 3.2555E+02 | 1.7747E+02 | 2.2516E+02 | 7.2408E+02 | 8.5310E+03 | 2.7604E+02 | 4.2926E+02 | 3.4070E+02 |
| F8 | Avg | 2.3153E+03 | 2.4193E+03 | 2.2310E+03 | 2.2324E+03 | 2.2224E+03 | 2.3537E+03 | 2.2290E+03 | 2.4171E+03 | 2.2224E+03 | 2.2241E+03 | 2.2211E+03 |
|  | Std | 1.1899E+03 | 2.7272E+04 | 8.9604E+02 | 8.7208E+02 | 1.2782E+00 | 1.8685E+05 | 5.3493E+01 | 1.8651E+04 | 1.0029E+00 | 2.7518E+01 | 1.1947E+01 |
| F9 | Avg | 2.4935E+03 | 2.4971E+03 | 2.4820E+03 | 2.5026E+03 | 2.4808E+03 | 2.6100E+03 | 2.4808E+03 | 2.4812E+03 | 2.4808E+03 | 2.4808E+03 | 2.4808E+03 |
|  | Std | 7.3402E+01 | 1.0619E+03 | 1.0095E+01 | 3.7589E+02 | 8.3300E-23 | 2.4678E+03 | 3.9300E-21 | 7.7460E-03 | 3.7200E-25 | 1.7400E-20 | 4.1200E-23 |
| F10 | Avg | 2.5011E+03 | 4.4796E+03 | 2.9458E+03 | 3.0890E+03 | 2.5051E+03 | 3.2196E+03 | 2.5260E+03 | 4.0394E+03 | 2.5070E+03 | 2.5133E+03 | 2.4855E+03 |
|  | Std | 4.9107E-02 | 8.0914E+05 | 2.3697E+05 | 2.0353E+05 | 6.7290E+02 | 1.2280E+06 | 4.1784E+03 | 7.5209E+05 | 2.9393E+03 | 2.7125E+03 | 1.2669E+03 |
| F11 | Avg | 3.7768E+03 | 3.3455E+03 | 3.0111E+03 | 3.5129E+03 | 2.9300E+03 | 5.5933E+03 | 2.9253E+03 | 3.4189E+03 | 2.9400E+03 | 2.9033E+03 | 2.9167E+03 |
|  | Std | 4.8404E+04 | 1.2588E+06 | 2.7757E+04 | 2.3169E+05 | 6.1000E+03 | 1.2507E+06 | 1.5425E+04 | 2.2891E+06 | 2.4000E+03 | 8.3222E+03 | 9.3889E+03 |
| F12 | Avg | 2.9770E+03 | 3.9106E+03 | 2.9807E+03 | 2.9639E+03 | 2.9595E+03 | 3.1601E+03 | 2.9750E+03 | 3.5943E+03 | 2.9581E+03 | 2.9528E+03 | 2.9571E+03 |
|  | Std | 3.1631E+02 | 8.5495E+04 | 1.4352E+03 | 3.4093E+02 | 3.6502E+02 | 1.4400E+04 | 1.3039E+03 | 3.0793E+04 | 2.5189E+02 | 1.1766E+02 | 1.7181E+02 |

Table 9. The Friedman test results based on the CEC2022 test functions in 10 and 20 dimensions.

| | GA | PSO | DE | GWO | COOT | RSO | GTO | AOA | IHAOAVOA | OAVOA | HWEAVOA |
|---|---|---|---|---|---|---|---|---|---|---|---|
| Dimension=10 | | | | | | | | | | | |
| F1 | 10 | 6 | 11 | 8 | 1 | 9 | 1 | 6 | 1 | 1 | 1 |
| F2 | 10 | 8 | 11 | 7 | 1 | 9 | 3 | 5 | 4 | 6 | 2 |
| F3 | 1 | 8 | 11 | 7 | 4 | 10 | 5 | 9 | 6 | 2 | 3 |
| F4 | 9 | 10 | 11 | 1 | 7 | 4 | 5 | 3 | 5 | 7 | 2 |
| F5 | 10 | 8 | 11 | 6 | 1 | 7 | 5 | 9 | 4 | 2 | 3 |
| F6 | 2 | 3 | 11 | 10 | 8 | 9 | 1 | 7 | 5 | 6 | 4 |
| F7 | 5 | 9 | 11 | 6 | 2 | 8 | 7 | 10 | 3 | 4 | 1 |
| F8 | 8 | 9 | 11 | 4 | 6 | 7 | 5 | 10 | 2 | 1 | 3 |
| F9 | 8 | 7 | 11 | 9 | 1 | 10 | 1 | 6 | 1 | 1 | 1 |
| F10 | 5 | 10 | 11 | 8 | 1 | 6 | 7 | 9 | 1 | 1 | 1 |
| F11 | 8 | 5 | 11 | 9 | 1 | 10 | 6 | 7 | 3 | 2 | 4 |
| F12 | 6 | 9 | 11 | 7 | 1 | 8 | 5 | 10 | 4 | 3 | 2 |
| Avg. Rank/No. | 6.83/6 | 7.67/9 | 11/11 | 6.83/6 | 2.83/2 | 8.08/10 | 4.25/5 | 7.58/8 | 3.25/4 | 3.00/3 | 2.25/1 |
| Dimension=20 | | | | | | | | | | | |
| F1 | 8 | 7 | 9 | 10 | 1 | 11 | 1 | 6 | 1 | 1 | 1 |
| F2 | 11 | 7 | 6 | 9 | 5 | 10 | 2 | 8 | 1 | 4 | 3 |
| F3 | 1 | 9 | 4 | 6 | 7 | 11 | 8 | 10 | 5 | 2 | 3 |
| F4 | 6 | 9 | 1 | 2 | 4 | 11 | 5 | 8 | 10 | 7 | 3 |
| F5 | 11 | 7 | 1 | 2 | 3 | 9 | 4 | 8 | 10 | 6 | 5 |
| F6 | 3 | 2 | 11 | 4 | 7 | 10 | 1 | 8 | 5 | 9 | 6 |
| F7 | 1 | 10 | 3 | 6 | 5 | 9 | 8 | 11 | 7 | 4 | 2 |
| F8 | 8 | 11 | 6 | 7 | 2 | 9 | 5 | 10 | 2 | 4 | 1 |
| F9 | 8 | 9 | 7 | 10 | 1 | 11 | 1 | 6 | 1 | 1 | 1 |
| F10 | 2 | 11 | 7 | 8 | 3 | 9 | 6 | 10 | 4 | 5 | 1 |
| F11 | 10 | 7 | 6 | 9 | 4 | 11 | 3 | 8 | 5 | 1 | 2 |
| F12 | 7 | 11 | 8 | 5 | 4 | 9 | 6 | 10 | 3 | 1 | 2 |
| Avg. Rank/No. | 6.33/7 | 8.33/9 | 5.75/6 | 6.50/8 | 3.83/4 | 10.00/11 | 4.17/3 | 8.58/10 | 4.50/5 | 3.75/2 | 2.50/1 |

Table 10. Sensitivity analysis of different population sizes based on 23 classical functions.

|   |   | 30 | 60 | 100 | 200 |
|---|---|---|---|---|---|
| F1 | Avg | 0.0000E+00 | 0.0000E+00 | 0.0000E+00 | 0.0000E+00 |
|    | Std | 0.0000E+00 | 0.0000E+00 | 0.0000E+00 | 0.0000E+00 |
| F2 | Avg | 0.0000E+00 | 0.0000E+00 | 0.0000E+00 | 0.0000E+00 |
|    | Std | 0.0000E+00 | 0.0000E+00 | 0.0000E+00 | 0.0000E+00 |
| F3 | Avg | 0.0000E+00 | 0.0000E+00 | 0.0000E+00 | 0.0000E+00 |
|    | Std | 0.0000E+00 | 0.0000E+00 | 0.0000E+00 | 0.0000E+00 |
| F4 | Avg | 0.0000E+00 | 0.0000E+00 | 0.0000E+00 | 0.0000E+00 |
|    | Std | 0.0000E+00 | 0.0000E+00 | 0.0000E+00 | 0.0000E+00 |
| F5 | Avg | 5.1640E-03 | 5.2367E-03 | 8.5621E-04 | 1.7253E-03 |
|    | Std | 9.3618E-06 | 3.4359E-05 | 1.4911E-07 | 9.4956E-06 |
| F5 | Avg | 5.5716E-07 | 1.1698E-06 | 1.7265E-06 | 1.2100E-07 |
|    | Std | 7.2768E-13 | 1.4465E-11 | 3.6144E-12 | 2.1064E-12 |
| F7 | Avg | 1.2076E-05 | 5.9481E-05 | 3.7508E-05 | 1.7748E-05 |
|    | Std | 1.4328E-09 | 3.5820E-09 | 1.5147E-09 | 3.0263E-10 |
| F8 | Avg | -1.2527E+04 | -1.2103E+04 | -1.2204E+04 | -1.2371E+04 |
|    | Std | 8.6230E+03 | 5.5886E+03 | 4.2403E+04 | 2.8936E+03 |
| F9 | Avg | 0.0000E+00 | 0.0000E+00 | 0.0000E+00 | 0.0000E+00 |
|    | Std | 0.0000E+00 | 0.0000E+00 | 0.0000E+00 | 0.0000E+00 |
| F10 | Avg | 8.8818E-16 | 8.8818E-16 | 8.8818E-16 | 8.8818E-16 |
|     | Std | 0.0000E+00 | 0.0000E+00 | 0.0000E+00 | 0.0000E+00 |
| F11 | Avg | 0.0000E+00 | 0.0000E+00 | 0.0000E+00 | 0.0000E+00 |
|     | Std | 0.0000E+00 | 0.0000E+00 | 0.0000E+00 | 0.0000E+00 |
| F12 | Avg | 5.7613E-09 | 2.5498E-07 | 2.5966E-08 | 2.6022E-08 |
|     | Std | 2.5338E-15 | 1.5462E-11 | 1.9803E-14 | 1.8059E-15 |
| F13 | Avg | 7.3758E-08 | 4.0968E-06 | 1.1280E-07 | 3.4556E-08 |
|     | Std | 1.1564E-14 | 5.0752E-12 | 4.2136E-14 | 3.7311E-15 |
| F14 | Avg | 1.2568E+00 | 1.4675E+00 | 1.2382E+00 | 1.0843E+00 |
|     | Std | 4.2014E-01 | 6.0182E-01 | 3.6571E-01 | 1.5693E-01 |
| F15 | Avg | 3.1114E-04 | 5.8826E-04 | 5.1128E-04 | 4.0008E-04 |
|     | Std | 5.9227E-09 | 6.2153E-09 | 5.8469E-09 | 2.7391E-08 |
| F16 | Avg | -1.0316E+00 | -1.0316E+00 | -1.0316E+00 | -1.0316E+00 |
|     | Std | 2.0454E-28 | 1.4192E-26 | 5.0793E-31 | 6.2861E-30 |
| F17 | Avg | 3.9789E-01 | 3.9789E-01 | 3.9789E-01 | 3.9789E-01 |
|     | Std | 5.4939E-28 | 5.3416E-29 | 3.9154E-26 | 6.7903E-29 |
| F18 | Avg | 3.0270E+00 | 3.0000E+00 | 3.0270E+00 | 3.0000E+00 |
|     | Std | 7.2827E-10 | 8.1005E-09 | 3.6158E-10 | 1.1593E-12 |
| F19 | Avg | -3.8628E+00 | -3.8628E+00 | -3.8628E+00 | -3.8628E+00 |
|     | Std | 8.6647E-15 | 4.5036E-18 | 1.7259E-20 | 1.6654E-15 |
| F20 | Avg | -3.2713E+00 | -3.2668E+00 | -3.2623E+00 | -3.2561E+00 |
|     | Std | 3.5391E-03 | 3.5244E-03 | 3.5389E-03 | 3.4928E-03 |
| F21 | Avg | -1.0153E+01 | -1.0153E+01 | -1.0153E+01 | -1.0153E+01 |
|     | Std | 1.9212E-20 | 5.1396E-21 | 7.5399E-18 | 4.7699E-20 |
| F22 | Avg | -1.0402E+01 | -1.0402E+01 | -1.0402E+01 | -1.0402E+01 |
|     | Std | 5.5748E-20 | 1.0652E-21 | 1.9089E-17 | 3.8890E-20 |
| F23 | Avg | -1.0536E+01 | -1.0536E+01 | -1.0536E+01 | -1.0536E+01 |
|     | Std | 3.9712E-20 | 7.5420E-21 | 1.7320E-17 | 7.6821E-20 |

## 4.6 Scalability analysis

In this section, the scalability of the proposed HWEAVOA and AVOA is investigated using the classical test functions with different dimension sizes, including dimension = 10, 30, 50, 100, and 1000. The experimental results of using thirteen classical functions (F1~F13) with different dimensions sizes is shown in Table 11.

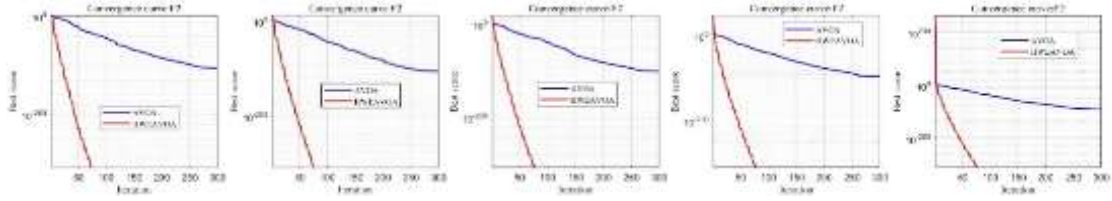

(1) the F2, dimension =10,30,50,100,1000

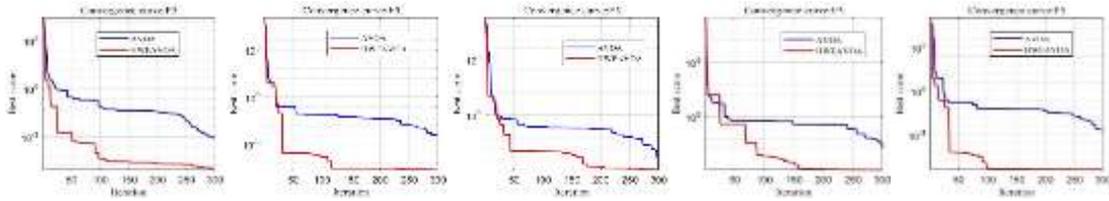

(2) the F5, dimension =10,30,50,100,1000

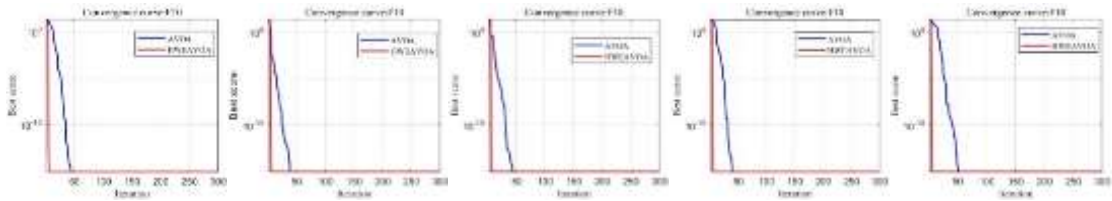

(3) the F10, dimension =10,30,50,100,1000

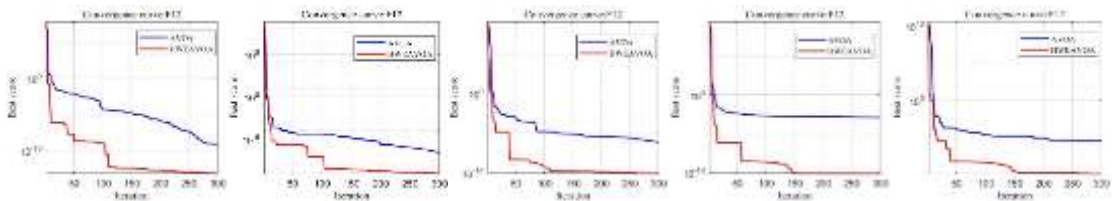

(4) the F12, dimension =10,30,50,100,1000

Fig. 7. Convergence curves of HWEAVOA and AVOA based on F2, F5, F10, F12 in different dimensions.

When solving functions F1~F4, F9, F10 and F11, the optimal values in five different function dimensions can be obtained by HWEAVOA. Although the optimal values of other functions in different dimensions cannot be obtained, the searchability and algorithm stability of HWEAVOA are still better than AVOA. As shown in Fig. 7, the HWEAVOA algorithm dramatically improves the convergence speed and solving accuracy in the solution space when dealing with problems in different dimensions compared with the original AVOA. Meanwhile, as shown in Table 6 and Table 7, the HWEAVOA all takes the leading positions in the top three when solving CEC2022 test functions F1~F4, F7~F10 and F12. The above analysis shows that HWEAVOA has a better ability to deal with high-dimensional problems than that of the original AVOA.

Table 11. Effects of different dimensions on HWEAVOA and AVOA.

| F1 | Avg | | Std | | F2 | Avg | | Std | |
|---|---|---|---|---|---|---|---|---|---|
| | HWEAVOA | AVOA | HWEAVOA | AVOA | | HWEAVOA | AVOA | HWEAVOA | AVOA |
| 10 | 0.0000E+00 | 7.4754E-156 | 0.0000E+00 | 4.1317E-308 | 10 | 0.0000E+00 | 3.8718E-85 | 0.0000E+00 | 9.6229E-167 |
| 30 | 0.0000E+00 | 1.8798E-152 | 0.0000E+00 | 3.4069E-301 | 30 | 0.0000E+00 | 5.5067E-79 | 0.0000E+00 | 2.9615E-154 |
| 50 | 0.0000E+00 | 2.5469E-155 | 0.0000E+00 | 4.8062E-307 | 50 | 0.0000E+00 | 2.8717E-80 | 0.0000E+00 | 8.0247E-157 |
| 100 | 0.0000E+00 | 7.5766E-155 | 0.0000E+00 | 5.2743E-306 | 100 | 0.0000E+00 | 7.7134E-76 | 0.0000E+00 | 5.9317E-148 |
| 1000 | 0.0000E+00 | 2.4051E-142 | 0.0000E+00 | 5.7688E-281 | 1000 | 0.0000E+00 | 3.7049E-89 | 0.0000E+00 | 1.1958E-174 |
| F3 | Avg | | Std | | F4 | Avg | | Std | |
| | HWEAVOA | AVOA | HWEAVOA | AVOA | | HWEAVOA | AVOA | HWEAVOA | AVOA |
| 10 | 0.0000E+00 | 6.6029E-126 | 0.0000E+00 | 4.3427E-248 | 10 | 0.0000E+00 | 8.3521E-82 | 0.0000E+00 | 3.5317E-160 |
| 30 | 0.0000E+00 | 6.8769E-109 | 0.0000E+00 | 4.6984E-214 | 30 | 0.0000E+00 | 8.8211E-77 | 0.0000E+00 | 6.4879E-150 |
| 50 | 0.0000E+00 | 2.4443E-99 | 0.0000E+00 | 5.9264E-195 | 50 | 0.0000E+00 | 4.8276E-78 | 0.0000E+00 | 1.3497E-152 |
| 100 | 0.0000E+00 | 4.8091E-86 | 0.0000E+00 | 2.3068E-168 | 100 | 0.0000E+00 | 3.5643E-76 | 0.0000E+00 | 1.2583E-148 |
| 1000 | 0.0000E+00 | 2.7765E-24 | 0.0000E+00 | 7.6591E-45 | 1000 | 0.0000E+00 | 3.7161E-78 | 0.0000E+00 | 1.3528E-152 |
| F5 | Avg | | Std | | F6 | Avg | | Std | |
| | HWEAVOA | AVOA | HWEAVOA | AVOA | | HWEAVOA | AVOA | HWEAVOA | AVOA |
| 10 | 1.5708E-01 | 1.0491E+00 | 5.6701E-01 | 3.2174E+00 | 10 | 1.1328E-09 | 1.7057E-05 | 2.1213E-18 | 2.6187E-07 |
| 30 | 8.3167E-02 | 1.6046E-01 | 2.2622E+00 | 4.2509E+00 | 30 | 2.5976E-04 | 1.2587E-03 | 4.2497E-06 | 3.0561E-04 |
| 50 | 4.7133E-02 | 2.1518E+00 | 9.5856E-02 | 4.4630E+00 | 50 | 1.5179E-03 | 3.2196E-04 | 2.0168E-03 | 4.6227E-04 |
| 100 | 2.5488E-03 | 1.1534E-05 | 3.8857E-03 | 3.6376E-04 | 100 | 4.8263E-03 | 2.2146E-02 | 1.2379E-03 | 6.1494E-03 |
| 1000 | 5.6482E-02 | 2.2465E-01 | 5.6786E-02 | 1.5965E+01 | 1000 | 6.9354E-02 | 4.9993E-01 | 2.6118E-01 | 1.0005E+00 |
| F7 | Avg | | Std | | F8 | Avg | | Std | |
| | HWEAVOA | AVOA | HWEAVOA | AVOA | | HWEAVOA | AVOA | HWEAVOA | AVOA |
| 10 | 1.1536E-04 | 2.7483E-04 | 1.2103E-08 | 8.9476E-08 | 10 | -3.9239E+03 | -3.8095E+05 | 1.6718E+05 | 2.6001E+05 |
| 30 | 1.1475E-04 | 2.8733E-04 | 1.1029E-08 | 9.6857E-08 | 30 | -1.2225E+04 | -1.2001E+04 | 3.8323E+05 | 7.7594E+05 |

| | Avg | | Std | | | Avg | | Std | |
|---|---|---|---|---|---|---|---|---|---|
| | HWEAVOA | AVOA | HWEAVOA | AVOA | | HWEAVOA | AVOA | HWEAVOA | AVOA |
| 50 | 1.2269E-04 | 2.7314E-04 | 1.5506E-08 | 8.6481E-08 | 50 | -2.0366E+04 | -1.9818E+04 | 9.7411E+05 | 2.4067E+06 |
| 100 | 1.1728E-04 | 2.8791E-04 | 1.4503E-08 | 9.6452E-08 | 100 | -4.0630E+04 | -3.8769E+04 | 4.1081E+06 | 1.3336E+07 |
| 1000 | 1.2293E-04 | 3.2956E-04 | 1.3847E-08 | 1.2399E-07 | 1000 | -4.0110E+05 | -3.8088E+05 | 5.5598E+08 | 1.8369E+09 |
| F9 | Avg | | Std | | F10 | Avg | | Std | |
| | HWEAVOA | AVOA | HWEAVOA | AVOA | | HWEAVOA | AVOA | HWEAVOA | AVOA |
| 10 | 0.0000E+00 | 0.0000E+00 | 0.0000E+00 | 0.0000E+00 | 10 | 8.8818E-16 | 8.8818E-16 | 0.0000E+00 | 0.0000E+00 |
| 30 | 0.0000E+00 | 0.0000E+00 | 0.0000E+00 | 0.0000E+00 | 30 | 8.8818E-16 | 8.8818E-16 | 0.0000E+00 | 0.0000E+00 |
| 50 | 0.0000E+00 | 0.0000E+00 | 0.0000E+00 | 0.0000E+00 | 50 | 8.8818E-16 | 8.8818E-16 | 0.0000E+00 | 0.0000E+00 |
| 100 | 0.0000E+00 | 0.0000E+00 | 0.0000E+00 | 0.0000E+00 | 100 | 8.8818E-16 | 8.8818E-16 | 0.0000E+00 | 0.0000E+00 |
| 1000 | 0.0000E+00 | 0.0000E+00 | 0.0000E+00 | 0.0000E+00 | 1000 | 8.8818E-16 | 8.8818E-16 | 0.0000E+00 | 0.0000E+00 |
| F11 | Avg | | Std | | F12 | Avg | | Std | |
| | HWEAVOA | AVOA | HWEAVOA | AVOA | | HWEAVOA | AVOA | HWEAVOA | AVOA |
| 10 | 0.0000E+00 | 0.0000E+00 | 0.0000E+00 | 0.0000E+00 | 10 | 1.9726E-09 | 1.1453E-04 | 1.6187E-17 | 1.8564E-06 |
| 30 | 0.0000E+00 | 0.0000E+00 | 0.0000E+00 | 0.0000E+00 | 30 | 4.4074E-05 | 5.4329E-05 | 1.6852E-07 | 5.4209E-07 |
| 50 | 0.0000E+00 | 0.0000E+00 | 0.0000E+00 | 0.0000E+00 | 50 | 5.9236E-05 | 8.2077E-05 | 1.8501E-07 | 4.2768E-07 |
| 100 | 0.0000E+00 | 0.0000E+00 | 0.0000E+00 | 0.0000E+00 | 100 | 4.5862E-05 | 1.1193E-04 | 7.3659E-08 | 3.2847E-07 |
| 1000 | 0.0000E+00 | 0.0000E+00 | 0.0000E+00 | 0.0000E+00 | 1000 | 1.8633E-05 | 8.8364E-05 | 2.0078E-08 | 8.0957E-08 |
| F13 | Avg | | Std | | | | | | |
| | HWEAVOA | AVOA | HWEAVOA | AVOA | | | | | |
| 10 | 1.1078E-05 | 1.4781E-04 | 1.2179E-07 | 1.2153E-05 | | | | | |
| 30 | 1.2236E-04 | 4.3281E-04 | 1.0019E-05 | 2.6475E-05 | | | | | |
| 50 | 2.3369E-04 | 2.7411E-04 | 1.3704E-05 | 2.0068E-05 | | | | | |
| 100 | 7.9216E-04 | 8.9577E-04 | 1.8364E-04 | 3.2531E-04 | | | | | |
| 1000 | 4.8569E-03 | 2.2227E-02 | 2.8346E-03 | 2.4359E-02 | | | | | |

**4.7 Time consumption and algorithm complexity analysis**

In order to study the time consumption of HWEAVOA, the running time of HWEAVOA and other advanced algorithms in classical test functions is shown in the Table 12, where the unit of measurement is seconds.

Experimental results show the time consumption of HWEAVOA is similar to that of the original AVOA and inferior compared to some classical optimization algorithms, but the algorithm accuracy and stability of HWEAVOA is leading. These experimental results demonstrate that the overall search performance of HWEAVOA has not been affected. Therefore, HWEAVOA has higher search efficiency than other advanced algorithms.

Table 12. Time consumption (measured in seconds) of different algorithms.

|     | AVOA   | GA     | PSO    | DE     | GWO    | COOT   | RSO    | GTO    | AOA    | IHAOAVOA | OAVOA  | HWEAVOA |
| --- | ------ | ------ | ------ | ------ | ------ | ------ | ------ | ------ | ------ | -------- | ------ | ------- |
| F1  | 0.0702 | 0.0347 | 0.0351 | 0.0029 | 0.1020 | 0.0353 | 0.0550 | 0.1174 | 0.0511 | 0.1316   | 0.1529 | 0.1061  |
| F2  | 0.0724 | 0.0351 | 0.0329 | 0.0029 | 0.0965 | 0.0370 | 0.0514 | 0.1229 | 0.0516 | 0.1345   | 0.1576 | 0.1126  |
| F3  | 0.1629 | 0.1213 | 0.2822 | 0.0049 | 0.1811 | 0.1183 | 0.1928 | 0.2837 | 0.1365 | 0.4085   | 0.3171 | 0.3728  |
| F4  | 0.0697 | 0.0334 | 0.0331 | 0.0028 | 0.0945 | 0.0355 | 0.0516 | 0.1172 | 0.0498 | 0.1255   | 0.1537 | 0.1025  |
| F5  | 0.0814 | 0.0425 | 0.0526 | 0.0031 | 0.1056 | 0.0453 | 0.0712 | 0.1412 | 0.0628 | 0.1631   | 0.1728 | 0.1345  |
| F6  | 0.0725 | 0.0333 | 0.0366 | 0.0028 | 0.0944 | 0.0360 | 0.0503 | 0.1180 | 0.0506 | 0.1261   | 0.1537 | 0.1022  |
| F7  | 0.1122 | 0.0706 | 0.1136 | 0.0041 | 0.1340 | 0.0793 | 0.1163 | 0.2000 | 0.0877 | 0.2398   | 0.2244 | 0.2177  |
| F8  | 0.0860 | 0.0437 | 0.0559 | 0.0031 | 0.1139 | 0.0483 | 0.0727 | 0.1393 | 0.0625 | 0.1565   | 0.1798 | 0.1245  |
| F9  | 0.0765 | 0.0404 | 0.0432 | 0.0031 | 0.0958 | 0.0418 | 0.0615 | 0.1211 | 0.0506 | 0.1308   | 0.1628 | 0.1038  |
| F10 | 0.0737 | 0.0388 | 0.0467 | 0.0031 | 0.0909 | 0.0399 | 0.0609 | 0.1135 | 0.0503 | 0.1223   | 0.1422 | 0.1029  |
| F11 | 0.0843 | 0.0496 | 0.0868 | 0.0033 | 0.1041 | 0.0511 | 0.0739 | 0.1370 | 0.0630 | 0.1591   | 0.1828 | 0.1304  |
| F12 | 0.1838 | 0.1509 | 0.3400 | 0.0061 | 0.2120 | 0.1533 | 0.2506 | 0.3218 | 0.1547 | 0.4169   | 0.4394 | 0.3328  |
| F13 | 0.1918 | 0.1505 | 0.3436 | 0.0060 | 0.2154 | 0.1545 | 0.2425 | 0.3390 | 0.1488 | 0.4317   | 0.4365 | 0.3406  |
| F14 | 0.2851 | 0.2215 | 0.4634 | 0.0069 | 0.2207 | 0.2173 | 0.4285 | 0.4383 | 0.1901 | 0.5863   | 0.7902 | 0.4102  |
| F15 | 0.0498 | 0.0180 | 0.0266 | 0.0011 | 0.0277 | 0.0287 | 0.0322 | 0.0820 | 0.0228 | 0.0985   | 0.1115 | 0.0786  |
| F16 | 0.0504 | 0.0165 | 0.0241 | 0.0009 | 0.0216 | 0.0270 | 0.0301 | 0.0790 | 0.0197 | 0.0917   | 0.1088 | 0.0755  |
| F17 | 0.0453 | 0.0191 | 0.0258 | 0.0009 | 0.0208 | 0.0291 | 0.1496 | 0.0799 | 0.0213 | 0.0955   | 0.1099 | 0.0693  |
| F18 | 0.0467 | 0.0129 | 0.0170 | 0.0008 | 0.0185 | 0.0245 | 0.0226 | 0.0743 | 0.0160 | 0.0790   | 0.0984 | 0.0635  |
| F19 | 0.0439 | 0.0177 | 0.0274 | 0.0011 | 0.0244 | 0.0268 | 0.0370 | 0.0752 | 0.0208 | 0.0851   | 0.0940 | 0.0750  |
| F20 | 0.0385 | 0.0161 | 0.0241 | 0.0013 | 0.0256 | 0.0227 | 0.0391 | 0.0645 | 0.0201 | 0.0755   | 0.0861 | 0.0633  |
| F21 | 0.0418 | 0.0193 | 0.0331 | 0.0014 | 0.026  | 0.0272 | 0.0462 | 0.0718 | 0.0230 | 0.0875   | 0.0936 | 0.0753  |
| F22 | 0.0550 | 0.0296 | 0.0510 | 0.0015 | 0.0370 | 0.0388 | 0.0558 | 0.1001 | 0.0322 | 0.1230   | 0.1283 | 0.1092  |
| F23 | 0.0756 | 0.0404 | 0.0719 | 0.0017 | 0.0475 | 0.0493 | 0.0685 | 0.1210 | 0.0421 | 0.1576   | 0.1473 | 0.1539  |

Moreover, the algorithm complexity is also an important evaluation basis for optimization algorithms. Therefore, the algorithm complexity is calculated based on the CEC2022 test functions. According to the Literature (Sun, Sun, Li, & Ieee, 2022), the calculation rules are as follows:

(1) Run the test program below:

$x = 0.55$

for $i = 1: 200000$

$x = x + x; x = x / 2; x = x * x, x = sqrt(x);$

$x = log(x); x = exp(x); x = x / (x + 2);$

end

(2) Evaluate the time consumption for CEC2022 test function F1 with 200,000 evaluations of a certain dimension $D$, it is $T1$;

(3) Evaluate the complete time for the mentioned algorithms with 200,000 evaluations of the same $D$ dimensional F1, it is $T2$;

(4) Calculate $T2$ for 5 independent runs, $T2 = mean(T2)$.

Finally, the algorithm complexity is calculated as $(T2 – T1) / T0$. According to the above method, the algorithm complexity for HWEAVOA and other improved algorithms is shown in Table 13.

Table 13. The algorithm complexity of the HWEAVOA and other advanced algorithms.

| Dimension | AVOA | GA | PSO | DE | GWO | COOT | RSO | GTO | AOA | IHAOAVOA | OAVOA | HWEAVOA |
|---|---|---|---|---|---|---|---|---|---|---|---|---|
| 10 | 72.83 | 45.81 | 82.86 | 2.29 | 60.98 | 54.50 | 58.49 | 132.43 | 54.34 | 193.54 | 163.92 | 74.24 |
| 20 | 72.70 | 47.30 | 80.51 | 2.37 | 71.00 | 53.48 | 59.12 | 134.24 | 57.68 | 190.63 | 171.96 | 73.54 |

Table 13 indicates that the HWEAVOA lost to part of the advanced algorithms for the algorithm complexity when Dimension = 10 and 20. However, HWEAVOA can achieve relatively competitive results with very small iterations throughout the convergence process, which lead other algorithms in the overall performance. On the whole, combined with the previous researches (Sun, Li, Huang, & Ieee, 2022), the result of the HWEAVOA has reached the general level of swarm intelligence algorithms and it is also competitive in the algorithm complexity.

**4.8 Wilcoxon rank sum test**

Although the test results of classical and CEC2022 test functions show the superiority of HWEAVOA in some extent, due to the stochastic nature of meta-heuristic algorithms, it is important to further illustrate the significant difference among these algorithms in the statistical way. The Wilcoxon rank sum test (Manalo, Biermann, Patil, & Mehta, 2022) can be counted the significant differences between algorithms, which is often used to evaluate the optimization performance of improved algorithms. Therefore, the Wilcoxon rank sum test is used on the CEC2022 test functions in 10 and 20 dimensions to further verify HWEAVOA at the level of $\alpha$ less than 0.05.

Table 14. *p*-Value for Wilcoxon rank and test results based on the CEC2022 test functions in 10 and 20 dimensions.

| | AVOA *p*-value win | GA *p*-value win | PSO *p*-value win | DE *p*-value win | GWO *p*-value win | COOT *p*-value win | RSO *p*-value win | GTO *p*-value win | AOA *p*-value win | IHAOAVOA *p*-value win | OAVOA *p*-value win |
|---|---|---|---|---|---|---|---|---|---|---|---|
| | | | | | | Dimension=10 | | | | | |
| F1 | 6.5183E-09 = | 2.8719E-11 + | 2.8719E-11 + | 1.0666E-07+ | 2.8719E-11 + | 2.8719E-11 = | 2.8719E-11 + | 7.9782E-02 = | 2.8719E-11 + | 2.1284E-09 = | 1.2685E-03 = |
| F2 | 3.1830E-01 + | 2.8719E-11 + | 2.8663E-02 + | 5.5611E-04+ | 4.2666E-06 + | 7.0069E-01 - | 2.8719E-11 + | 6.0484E-01 + | 1.0707E-01 + | 3.6322E-01 + | 8.7663E-01 + |
| F3 | 6.8023E-08 + | 1.9494E-02 - | 2.8719E-11 + | 1.3825E-05+ | 2.5495E-01 + | 8.7945E-04 + | 2.8719E-11 + | 8.5825E-06 + | 2.8719E-11 + | 1.2470E-02 + | 1.4787E-05 - |
| F4 | 1.2597E-01 + | 4.7792E-01 + | 5.3464E-01 + | 5.5727E-10+ | 5.8188E-07 - | 4.7885E-05 + | 6.0361E-04 + | 1.6461E-01 + | 3.2918E-01 + | 1.5798E-01 + | 6.8432E-01 + |
| F5 | 3.3681E-05 - | 1.7938E-06 + | 1.0727E-04 + | 3.3258E-02+ | 3.2054E-02 + | 1.3690E-08 - | 2.6046E-04 + | 3.3258E-02 + | 4.7885E-05 + | 8.8247E-01 + | 4.4205E-06 - |
| F6 | 6.6763E-02 + | 2.8719E-11 - | 1.1513E-08 - | 3.0566E-05+ | 3.9739E-06 + | 9.6739E-03 + | 1.2556E-08 + | 2.8719E-11 - | 6.8990E-02 + | 9.5284E-01 + | 9.4107E-01 + |
| F7 | 1.1284E-05 + | 5.7668E-08 + | 3.8787E-11 + | 1.4119E-02+ | 4.1614E-01 + | 1.3549E-02 + | 2.8719E-11 + | 6.5216E-03 + | 2.8719E-11 + | 1.4493E-04 + | 1.9494E-02 + |
| F8 | 9.5284E-01 + | 2.8719E-11 + | 5.7730E-11 + | 3.0071E-01+ | 1.0856E-03 + | 1.3549E-02 + | 2.8719E-11 + | 4.7454E-03 + | 2.8719E-11 + | 3.9117E-01 - | 2.8047E-01 - |
| F9 | 1.2658E-10 = | 2.8719E-11 + | 1.2658E-10 + | 3.9395E-03+ | 2.8719E-11 + | 2.8719E-11 = | 3.9998E-09 + | 2.8719E-11 = | 2.8719E-11 + | 2.8719E-11 = | 8.2450E-01 = |
| F10 | 3.3656E-01 + | 1.4376E-06 + | 3.0561E-09 + | 1.4787E-05+ | 1.0097E-02 + | 1.2044E-07 = | 1.0670E-06 + | 1.4796E-03 + | 5.3167E-10 + | 3.5783E-02 = | 1.1959E-02 = |
| F11 | 1.1467E-02 + | 7.8929E-07 + | 1.2077E-05 + | 7.7863E-03+ | 3.7006E-06 + | 3.8778E-04 - | 1.6266E-08 + | 2.1105E-07 + | 9.1932E-06 + | 9.1809E-07 - | 1.0856E-03 - |
| F12 | 9.3341E-02 + | 8.1211E-09 + | 3.8787E-11 + | 3.9881E-04+ | 1.6016E-01 + | 7.7863E-03 - | 2.7927E-09 + | 8.3026E-01 + | 2.8719E-11 + | 5.4441E-01 + | 2.4625E-02 + |
| +/=/- | 9/2/1 | 10/0/2 | 11/0/1 | 12/0/0 | 11/0/1 | 5/3/4 | 12/0/0 | 9/2/1 | 12/0/0 | 7/3/2 | 5/3/4 |
| | | | | | | Dimension=20 | | | | | |
| F1 | 8.1014E-10 + | 2.8719E-11 + | 2.8719E-11 + | 2.8719E-11 + | 2.8719E-11 + | 2.8719E-11 = | 2.8719E-11 + | 7.2208E-06 = | 2.8719E-11 + | 2.6099E-10 = | 5.2650E-05 = |
| F2 | 1.6238E-01 + | 2.8719E-11 + | 1.0541E-05 + | 5.7460E-02 + | 6.4146E-10 + | 1.9494E-02 + | 2.8719E-11 + | 7.3383E-05 - | 2.4879E-10 + | 3.6709E-03 - | 5.2014E-04 + |
| F3 | 9.1809E-07 - | 4.2855E-11 + | 2.8719E-11 + | 4.9149E-06 + | 2.9757E-02 + | 3.5164E-01 + | 2.8719E-11 + | 3.8787E-11 + | 2.8719E-11 + | 2.7587E-04 + | 5.3081E-08 - |
| F4 | 6.1520E-02 + | 8.9092E-02 + | 3.6714E-01 - | 1.3950E-10 - | 3.5098E-11 + | 2.4726E-07 + | 1.7739E-09 + | 4.7454E-03 + | 2.3691E-01 + | 4.3764E-01 + | 1.5367E-01 + |
| F5 | 5.4441E-04 + | 6.5575E-05 + | 4.1614E-01 - | 2.8719E-11 - | 2.8719E-11 - | 3.8787E-11 + | 9.2932E-10 - | 7.4937E-04 + | 9.7641E-01 + | 7.3628E-02 + | 2.5495E-01 + |
| F6 | 3.3614E-01 - | 2.8719E-11 - | 7.8991E-05 + | 7.3628E-02 - | 5.4441E-10 + | 9.1757E-03 + | 2.8719E-11 + | 8.0170E-08 - | 1.7378E-01 + | 5.4609E-02 - | 3.6714E-01 + |
| F7 | 4.8713E-01 - | 1.2046E-03 + | 7.0436E-10 + | 2.6750E-01 + | 3.9117E-01 + | 1.7800E-07 + | 5.3167E-10 + | 2.3542E-05 + | 1.5370E-10 + | 2.8719E-11 + | 3.1752E-11 + |
| F8 | 2.8711E-01 + | 2.8719E-11 + | 5.2283E-11 + | 2.8711E-01 + | 1.0727E-04 + | 4.2820E-02 + | 5.2283E-11 + | 1.6906E-05 + | 5.2283E-11 + | 2.2798E-02 + | 1.5581E-01 + |
| F9 | 3.1752E-11 + | 2.8719E-11 + | 2.8719E-11 + | 2.8719E-11 + | 2.8719E-11 + | 2.3768E-07 = | 2.8719E-11 + | 2.2539E-01 = | 2.8719E-11 + | 3.1752E-11 = | 2.8719E-11 = |
| F10 | 1.8824E-01 + | 1.0241E-07 + | 5.2283E-11 + | 9.3135E-10 + | 7.7326E-10 + | 1.5581E-01 + | 3.5098E-11 + | 1.4153E-07 + | 1.2658E-10 + | 3.5933E-01 + | 8.5918E-04 + |
| F11 | 1.1436E-03 + | 2.8719E-11 + | 2.5138E-05 + | 3.6658E-04 + | 8.5630E-11 + | 6.3735E-04 + | 2.8719E-11 + | 1.6378E-03 + | 7.1014E-04 + | 5.9613E-03 + | 5.4336E-05 - |
| F12 | 3.5164E-01 + | 3.9395E-03 + | 2.8719E-11 + | 6.8990E-02 + | 1.5323E-02 + | 7.8519E-02 + | 2.8719E-11 + | 8.3602E-03 + | 2.8719E-11 + | 8.3670E-02 + | 2.2106E-03 - |
| +/=/- | 9/0/3 | 11/0/1 | 10/0/2 | 9/0/3 | 11/0/1 | 10/2/0 | 11/0/1 | 8/2/2 | 12/0/0 | 8/2/2 | 7/2/3 |

The statistical results of the Wilcoxon rank sum test of the HWEAVOA relative to each comparison algorithm are shown in Table 14, which "+", "=" and "-" is used to respectively indicate that HWEAVOA is superior or uniform or worse than the comparison algorithms. The *p*-value less than 0.05 indicates that HWEAVOA shows a more significant difference than the compared algorithm.

According to the test results in Table 13, compared with other advanced algorithms, the HWEAVOA's test *p*-values are all less than 0.05, and most symbols are "+". Therefore, HWEAVOA is statistically superior to other advanced algorithms, and there are significant differences between HWEAVOA and these algorithms.

## 5. Conclusions and future works

To address the issues of the original AVOA (i.e., the tendency to fall into local optimum and unbalance of global search stage and local search stage), an improved African vultures optimization algorithm (HWEAVOA) is proposed with three efficient optimization strategies. Firstly, the Henon chaotic mapping theory and elite population strategy are proposed, which improve the vulture initial population's randomness and diversity; Furthermore, the nonlinear adaptive incremental inertial weight factor is introduced in the location update phase, which satisfies the requirement for the exploration and exploitation ability in different phases, and avoid the algorithm falling into a local optimum; The addition of the reverse learning competition strategy allows the algorithm to expand the discovery fields for the optimal solution, accelerate the convergence speed of the algorithm and strengthen the ability to jump out of the local optimal solution. In terms of simulation test, HWEAVOA and other advanced comparison algorithms are used to solve classical and CEC2022 test functions. Through the comparative analysis of experimental results and convergence curves, it is proved that the optimization ability and convergence speed of the HWEAVOA for solving complex functions are obviously better than the other advanced algorithms. Meanwhile, HWEAVOA has reached the general level in the algorithm complexity, and its overall performance is competitive in the swarm intelligence algorithms.

The future possible works are as follows:

Although the proposed HWEAVOA improves the algorithm performance in the optimization ability, convergence speed and solution stability, it found that the HWEAVOA also has the room for improvement in time consumption and algorithmic complexity according to the argument in the article. Follow-up work will be further optimized the HWEAVOA algorithm for this issue. Furthermore, large-scale problems and dynamic problems are the current development trend of the swarm intelligence, but many algorithms perform poorly and are prone to local optimization when solving these problems. We will continue to study on the base of the HWEAVOA, and improve the applied space of swarm intelligent algorithms.

## CRediT authorship contribution statement

**Baiyi Wang:** Conceptualization, Methodology, Analysis, Data Curation, Writing - review & editing, Funding acquisition. **Zipeng Zhang:** Methodology, Supervision, Data collection, Writing – original draft. **Patrick Siarry:** Investigation, Data collection, Writing – review & editing. **Xinhua Liu:** Supervision, Funding acquisition. **Grzegorz Królczyk:** Analysis, Writing – review & editing. **Dezheng Hua:** Analysis, Writing - Review & Editing. **Frantisek Brumercik:** Data collection, Visualization. **Zhixiong Li:** Investigation, Project administration, Writing - review & editing.

## Declaration of Competing Interest

The authors declare that they have no known competing financial interests or personal relationships that could have appeared to influence the work reported in this paper.

## Data Availability

All data that produce the results in this work can be requested from the corresponding author.

## Acknowledgements

The support of National Natural Science Foundation of China (No. 51975568), the Independent Innovation Project of "Double-First Class" Construction of China University of Mining and Technology (2022ZZCX06) and Postgraduate Research & Practice Innovation Program of Jiangsu Province (KYCX23_2681) in carrying out this research are gratefully acknowledged.

355 **Appendix 1**

| Unimodal benchmark functions | | | |
|---|---|---|---|
| Function equation | Dim | Range | Optimal |
| F1 $f_1(x) = \sum_{i=1}^{n} x_i^2$ | 30 | [-100,100] | 0 |
| F2 $f_2(x) = \sum_{i=1}^{n}|x_i| + \prod_{i=1}^{n}|x_i|$ | 30 | [-10,10] | 0 |
| F3 $f_3(x) = \sum_{i=1}^{n}\left(\sum_{j=1}^{i} x_j\right)^2$ | 30 | [-100,100] | 0 |
| F4 $f_4(x) = \max_i\{|x_i|, 1 \leq i \leq n\}$ | 30 | [-100,100] | 0 |
| F5 $f_5(x) = \sum_{i=1}^{n-1}\left[100(x_{i+1} - x_i^2)^2 + (x_i - 1)^2\right]$ | 30 | [-30,30] | 0 |
| F6 $f_6(x) = \sum_{i=1}^{n}([x_i + 0.5])^2$ | 30 | [-100,100] | 0 |
| F7 $f_7(x) = \sum_{i=1}^{n} i x_i^4 + random[0,1)$ | 30 | [-1.28,1.28] | 0 |

| Multimodal benchmark functions | | | |
|---|---|---|---|
| Function equation | Dim | Range | Optimal |
| F8 $f_8(x) = \sum_{i=1}^{n} -x_i \sin\left(\sqrt{|x_i|}\right)$ | 30 | [-500,500] | $-418.9829 \times Dim$ |
| F9 $f_9(x) = \sum_{i=1}^{n}\left[x_i^2 - 10\cos(2\pi x_i) + 10\right]$ | 30 | [-5.12,5.12] | 0 |
| F10 $f_{10}(x) = -20\exp\left(-0.2\sqrt{\frac{1}{n}\sum_{i=1}^{n} x^i}\right) - \exp\left(\frac{1}{n}\sum_{i=1}^{n}\cos(2\pi x_i)\right) + 20 + e$ | 30 | [-32,32] | 0 |
| F11 $f_{11}(x) = \frac{1}{4000}\sum_{i=1}^{n} x_i^2 - \prod_{i=1}^{n}\cos\left(\frac{x_i}{\sqrt{i}}\right) + 1$ | 30 | [-600,600] | 0 |
| F12 $f_{12}(x) = \frac{\pi}{n}10\sin(\pi y_1) + \sum_{i=1}^{n-1}\left\{(y_i - 1)^2 \times \left[1 + 10\sin^2(\pi y_{i+1})\right] + \sum_{i=1}^{n}\mu(x_i, 10, 100, 4)\right\}$, where $y_i = 1 + \frac{x_i + 1}{4}$, | 30 | [-50,50] | 0 |

$$\mu(x_i, a, k, m) = \begin{cases} k(x_i - a)^m, & x_i > a \\ 0, & -a < x_i < a \\ k(-x_i - a)^m, & x_i < -a \end{cases}$$

| | Function equation | Dim | Range | Optimal |
|---|---|---|---|---|
| F13 | $f_{13}(x) = 0.1\left\{\sin^2(3\pi x_1) + \sum_{i=1}^{n}(x_i - 1)^2\left[1 + \sin^2(3\pi x_i + 1)\right]\right.$ $\left. + (x_n - 1)^2[1 + \sin^2(2\pi x_n)]\right\} + \sum_{i=1}^{n}\mu(x_i, 5, 100, 4)$ | 30 | [-50,50] | 0 |
| **Fixed-dimension multimodal benchmark functions** | | | | |
| | Function equation | Dim | Range | Optimal |
| F14 | $f_{14}(x) = \left(\dfrac{1}{500} + \sum_{j=1}^{25}\dfrac{1}{j + \sum_{i=1}^{2}(x_i - a_{ij})^6}\right)$ | 30 | [-65,65] | 1 |
| F15 | $f_{15}(x) = \sum_{i=1}^{11}\left[a_i - \dfrac{x_1(b_i^2 + b_i x_2)}{b_i^2 + b_i x_3 + x_4}\right]^2$ | 30 | [-5,5] | 0.00030 |
| F16 | $f_{16}(x) = 4x_1^2 - 2.1x_1^4 + \dfrac{1}{3}x_1^6 + x_1 x_2 - 4x_2^2 + 4x_2^4$ | 30 | [-5,5] | -1.0316 |
| F17 | $f_{17}(x) = \left(x_2 - \dfrac{5.1}{4\pi^2}x_1^2 + \dfrac{5}{\pi}x_1 - 6\right)^2 + 10\left(1 - \dfrac{1}{8\pi}\right)\cos x_1 + 10$ | 30 | [-5,5] | 0.398 |
| F18 | $f_{18}(x) = \left[1 + (x_1 + x_2 + 1)^2(19 - 14x_1 + 3x_1^2 - 14x_2 + 6x_1 x_2 + 3x_2^2)\right] \times$ $\left[30 + (2x_1 - 3x_2)^2 \times (18 - 32x_1 + 12x_1^2 + 48x_2 - 36x_1 x_2 + 27x_2^2)\right]$ | 30 | [-2,2] | 3 |
| F19 | $f_{19}(x) = -\sum_{i=1}^{4}c_i \exp\left(-\sum_{j=1}^{3}a_{ij}(x_i - p_{ij})^2\right)$ | 30 | [1,3] | -3.86 |
| F20 | $f_{20}(x) = -\sum_{i=1}^{4}c_i \exp\left(-\sum_{j=1}^{6}a_{ij}(x_i - p_{ij})^2\right)$ | 30 | [0,1] | -3.32 |
| F21 | $f_{21}(x) = -\sum_{i=1}^{5}\left[(X - a_i)(X - a_i)^T + c_i\right]^{-1}$ | 30 | [0,10] | -10.1532 |
| F22 | $f_{22}(x) = -\sum_{i=1}^{7}\left[(X - a_i)(X - a_i)^T + c_i\right]^{-1}$ | 30 | [0,10] | -10.4028 |
| F23 | $f_{23}(x) = -\sum_{i=1}^{10}\left[(X - a_i)(X - a_i)^T + c_i\right]^{-1}$ | 30 | [0,10] | -10.5363 |

356

357 # Appendix 2

|  | Function | Functions | $F_i^*$ |
|---|---|---|---|
| Unimodal Function | F1 | Shifted and full Rotated Zakharov Function | 300 |
| Basic Functions | F2 | Shifted and full Rotated Rosenbrock's Function | 400 |
|  | F3 | Shifted and full Rotated Expanded Schaffer's $f6$ Function | 600 |
|  | F4 | Shifted and full Rotated Non-Continuous Rastrigin's Function | 800 |
|  | F5 | Shifted and full Rotated Levy Function | 900 |
| Hybrid Functions | F6 | Hybrid Function 1 (N = 3) | 1800 |
|  | F7 | Hybrid Function 2 (N = 6) | 2000 |
|  | F8 | Hybrid Function 3 (N = 5) | 2200 |
| Composition Functions | F9 | Composition Function 1 (N = 5) | 2300 |
|  | F10 | Composition Function 2 (N = 4) | 2400 |
|  | F11 | Composition Function 3 (N = 5) | 2600 |
|  | F12 | Composition Function 3 (N = 6) | 2700 |
| Search range: $[-100, 100]^D$ | | | |

358